\let\accentvec\vec

\documentclass[runningheads]{llncs}

\let\vec\accentvec
\usepackage{graphicx}
\usepackage{comment}
\usepackage{amsfonts,amsmath,amssymb} % define this before the line numbering.
\usepackage{color}
\usepackage{mathrsfs}
\usepackage{subfigure}
\usepackage{multirow}
\usepackage{caption2}
\usepackage[misc]{ifsym}

\DeclareMathOperator*{\argmin}{arg\,min}

\usepackage{xspace}
\makeatletter
\DeclareRobustCommand\onedot{\futurelet\@let@token\@onedot}
\def\@onedot{\ifx\@let@token.\else.\null\fi\xspace}

\def\etal{\emph{et al}\onedot}
\makeatother
\begin{document}

\setlength{\abovedisplayskip}{3pt}
\setlength{\belowdisplayskip}{3pt}

% \renewcommand\makeLineNumber {\hss\thelinenumber\ \hspace{6mm} \rlap{\hskip\textwidth\ \%hspace{6.5mm}\thelinenumber}}
% \linenumbers
\pagestyle{headings}
\mainmatter
\def\ECCVSubNumber{3112}  % Insert your submission number here

\title{Improving Deep Video Compression by Resolution-adaptive Flow Coding} % Replace with your title
% \begin{comment}
% \titlerunning{ECCV-20 submission ID \ECCVSubNumber} 
% \authorrunning{ECCV-20 submission ID \ECCVSubNumber} 
% \author{Anonymous ECCV submission}
% \institute{Paper ID \ECCVSubNumber}
% \end{comment}
%******************

% \begin{comment}
\titlerunning{Improving Deep Video Compression by Resolution-adaptive Flow Coding}
\author{Zhihao Hu\inst{1*} \and 
Zhenghao Chen\inst{2*} \and
Dong Xu\inst{2} \and \\
\Letter~Guo Lu\inst{3} \and
Wanli Ouyang\inst{2} \and
Shuhang Gu\inst{2}}
\authorrunning{Zhihao Hu et al.}
\institute{College of Software, Beihang University, China\\
% \email{huzhihao@buaa.edu.cn} 
\and
School of Electrical and Information Engineering,\\ The University of Sydney, Australia \\
% \email{\{zhenghao.chen,dong.xu,wanli.ouyang\}@sydney.edu.au, shuhanggu@gmail.com} 
\and
School of Computer Science \& Technology, Beijing Institute of Technology, China \\
% \email{sdluguo@gmail.com}
}
% \end{comment}
%******************
\maketitle
\let\thefootnote\relax\footnotetext{*First two authors contributed equally.\\
\Letter~Corresponding author: Guo Lu (sdluguo@gmail.com).}
\begin{abstract}

In the learning based video compression approaches, it is an essential issue to compress pixel-level optical flow maps by developing new motion vector (MV) encoders. In this work, we propose a new framework called Resolution-adaptive Flow Coding (RaFC) to effectively compress the flow maps globally and locally, in which we use multi-resolution representations instead of single-resolution representations for both the input flow maps and the output motion features of the MV encoder. To handle complex or simple motion patterns globally, our frame-level scheme RaFC-frame automatically decides the optimal flow map resolution for each video frame. To cope different types of motion patterns locally, our block-level scheme called RaFC-block can also select the optimal resolution for each local block of motion features. In addition, the rate-distortion criterion is applied to both RaFC-frame and RaFC-block and select the optimal motion coding mode for effective flow coding. Comprehensive experiments on four benchmark datasets HEVC, VTL, UVG and MCL-JCV clearly demonstrate the effectiveness of our overall RaFC framework after combing RaFC-frame and RaFC-block for video compression.
\end{abstract}

\section{Introduction}

There is increasing demand for new video compression systems to effectively reduce redundancy in video sequences.  The conventional video compression systems are based on hand-designed modules such as block based motion estimation and Discrete Cosine Transform (DCT). Taking advantage of large-scale training datasets and powerful nonlinear modeling capacity of deep neural networks, the recent deep video compression methods~\cite{lu2019dvc,wu2018video,tsai2018learning} have achieved promising video compression performance (Please refer to Section 2 for more details about the related image and video compression methods). Specifically, in the recent end-to-end deep video compression (DVC) framework~\cite{lu2019dvc}, all modules (e.g., DCT, motion estimation and motion compensation) in the conventional H.264/H.265 codec are replaced with the well-designed neural networks. 

In the learning based video compression approaches such as the aforementioned DVC framework, it is a non-trivial task to compress pixel-level optical flow maps. However, such frameworks adopt single representations for both input flow maps and output motion features using a single motion vector (MV) encoder. This cannot effectively handle complex or simple motion patterns in different scenes and fast or slow movement of objects. To this end, in this work we propose a new framework called Resolution-adaptive Flow Coding (RaFC),
%. This framework 
which can adopt multi-resolution representations for both flow maps and motion features and then automatically decide the optimal resolutions at both frame-level and block-level in order to achieve the optimal rate-distortion trade-off. 

At the frame-level, our RaFC-frame scheme can automatically decide the optimal flow map resolution for each video frame in order to effectively handle complex or simple motion patterns globally. As a result, for those frames with complex global motion patterns, high-resolution flow maps containing more detailed optimal flow information are more likely to be selected as the input for the MV encoder. In contrast, for the frames with simple global motion patterns, low-resolution optimal flow maps are generally preferred.
%to obtain the optimal rate-distortion value.
% in order to achieve the optimal rate-distortion trade-off. 

Inspired by the traditional codecs~\cite{wiegand2003overview,sullivan2012overview}, in which the blocks with different sizes are used for motion estimation, we also propose a new scheme RaFC-block, which can decide the optimal resolution for each block based on the rate-distortion (RD) criterion when encoding the motion features. As a result, for the local blocks with complicated motion patterns, our RaFC-block scheme will use high-resolution blocks containing fine motion features. For the blocks within smooth areas, our RaFC-block scheme prefers low-resolution blocks with coarse motion features in order to save bits for encoding their motion features without substantially sacrificing the distortion. In addition, we also propose an overall RaFC framework by combining the two newly proposed schemes RaFC-frame and RaFC-block. 

% Traditional video compression system always use the rate-distortion optimization (RDO) technique to select the optimal mode for compression. So the RDO technique is applied in our approach to automatically decide the optimal mode for both RaFC-frame and RaFC-block.
% As far as we know, we are the first learning based video compression method that uses the RD criterion to achieve high efficiency flow coding. 

We perform comprehensive experiments on four benchmark datasets HEVC Class E, VTL, UVG and MCL-JCV. The results clearly demonstrate our overall RaFC framework outperforms the baseline algorithms including H.264, H.265 and DVC. 
% In addition, the results of our RaFC framework can be further improved by adopting the latest image compression technology~\cite{minnen2018joint} to compress the residual information. 
Our contributions are summarized as follows:

% \{-2mm}

\begin{itemize}
  \item  To effectively handle complex or simple motion patterns globally, we adopt the multi-resolution representations for the flow maps, in which the optimal resolution at the frame-level can be automatically decided for our method RaFC-Frame based on the RD criterion. 

  \item Using multi-resolution representations for motion features, we additionally propose the RaFC-block method to automatically decide the optimal resolution at the block-level based on the RD criterion, which can effectively cope with different types of local motion patterns.
  
%   \item To automatically select the optimal resolution at both frame-level and block-level, we apply the rate-distortion optimization (RDO) technique in the learning based video compression for high efficiency flow coding.

  \item Our overall RaFC framework after combining RaFC-frame and RaFC-block achieves the state-of-the-arts video compression performance on four benchmark datasets including HEVC Class E, VTL, UVG and MCL-JCV.
  
\end{itemize}
%------------------------------------------------------------------------
\section{Related Work}
\subsection{Image Compression}
% Transform based approaches 

Transform-based image compression methods can efficiently reduce the spatial redundancy.
Currently, those approaches (e.g., JPEG~\cite{wallace1992jpeg}, BPG~\cite{bellard2015bpg} and JPEG2000~\cite{taubman2002jpeg2000}) are still the most widely used image compression algorithms.  
Recently, the deep learning based image compression methods~\cite{toderici2015variable,toderici2017full,balle2016end,balle2018variational,theis2017lossy,johnston2018improved,agustsson2017soft,li2018learning,rippel2017real,agustsson2019generative} have been proposed and achieved the state-of-the-arts performance.
The general idea of deep image compression is to transform input images into quantized bit-streams, which can be further compressed through lossless coding algorithms.
To achieve this goal,
some methods~\cite{toderici2017full,johnston2018improved,toderici2015variable} directly employed recurrent neural networks (RNNs) to compress the images in a progressive manner.
Toderici \etal~\cite{toderici2015variable} firstly introduced a simple RNN-based approach to compress the image and further proposed a method~\cite{toderici2017full}, which enhances the performance by progressively compressing reconstructed residual information. Johnston \etal~\cite{johnston2018improved} also improved Toderici's work by introducing a new objective loss.
Other popular approaches use an auto-encoder architecture~\cite{balle2016end,balle2018variational,minnen2018joint,theis2017lossy}.
%
% Specifically, 
Balle \etal~\cite{balle2016end} introduced a continuous and differentiable proxy for the rate-distortion loss and further proposed a variational auto-encoder based compression algorithm~\cite{balle2018variational}.

Recently, some methods~\cite{balle2018variational,minnen2018joint} focus on predicting different distribution in different spatial area. And Li \etal~\cite{li2018learning} introduced the importance map to reduce the total binary codes to transmit. All such methods need to transmit the full-resolution feature map to the decoding stage. Our proposed method selects the most optimal resolution at both frame-level and block-level in the encoding side, which saves a lot of bits.

% \{-5mm}
\subsection{Video Compression}

Traditional video compression algorithms, such as H.264~\cite{wiegand2003overview} and H.265~\cite{sullivan2012overview},  adopted the hand-crafted operations for motion estimation and motion compensation for inter-frame prediction. Even though they can successfully reduce temporal redundancy of video data, those compression algorithms are limited in compression performance as they cannot be jointly optimized.

With the success of deep learning based motion estimation and image compression approaches, some attempts have been made to use neural networks for video compression~\cite{tsai2018learning,wu2018video,chen2019learning,xu2018reduction}, in which the neural networks are used to replace the modules from the conventional approach.
The work in~\cite{chen2019learning} proposed a block based approach, while
Tsai \etal~\cite{tsai2018learning} utilized an auto-encoder approach to compress  residual information from H.264. 
Wu \etal~\cite{wu2018video} predicted and reconstructed video frames by using interpolation.
While the above works have achieved remarkable performance, they cannot be trained in an end-to-end fashion, which limits their performance.

Recently, more deep video compression methods~\cite{lu2019dvc,lu2020anendtoend,guo2020content,rippel2019learned,abdelaziz2019neural} have been proposed.
Lu \etal~\cite{lu2019dvc} proposed the first end-to-end deep learning video compression (DVC) framework, which replaces all the key components of the traditional video compression codec with deep neural networks.
Rippel \etal~\cite{rippel2019learned} proposed to maintain a state, which contains the past information, compressed motion information and residual information for video compression.
Djelouah \etal~\cite{abdelaziz2019neural} proposed an interpolation based video compression approach, which combines motion compression and image synthesis in a single network.
In these works, optical flow information plays an essential role.
In order to achieve reasonable compression performance, the state-of-the-art optical flow estimation networks~\cite{dosovitskiy2015flownet,hui2018liteflownet} have been adopted  
to provide accurate motion estimation.
However, as these optical flow estimation networks were designed for generating accurate full-resolution motion maps, they are not optimal for the video compression task.
Recently, Habibian \etal~\cite{habibian2019video} proposed a 3D auto-encoder approach without requiring  optical flow for motion compensation.
However, 
their algorithm is still limited for capturing fine scale motions. 

In contrast to these works, we propose a new framework RaFC to effectively compress optical flow maps, and 
% our network
it can be trained in an end-to-end fashion.

\begin{figure}[!t]
% \begin{center}
\includegraphics[width=\linewidth]{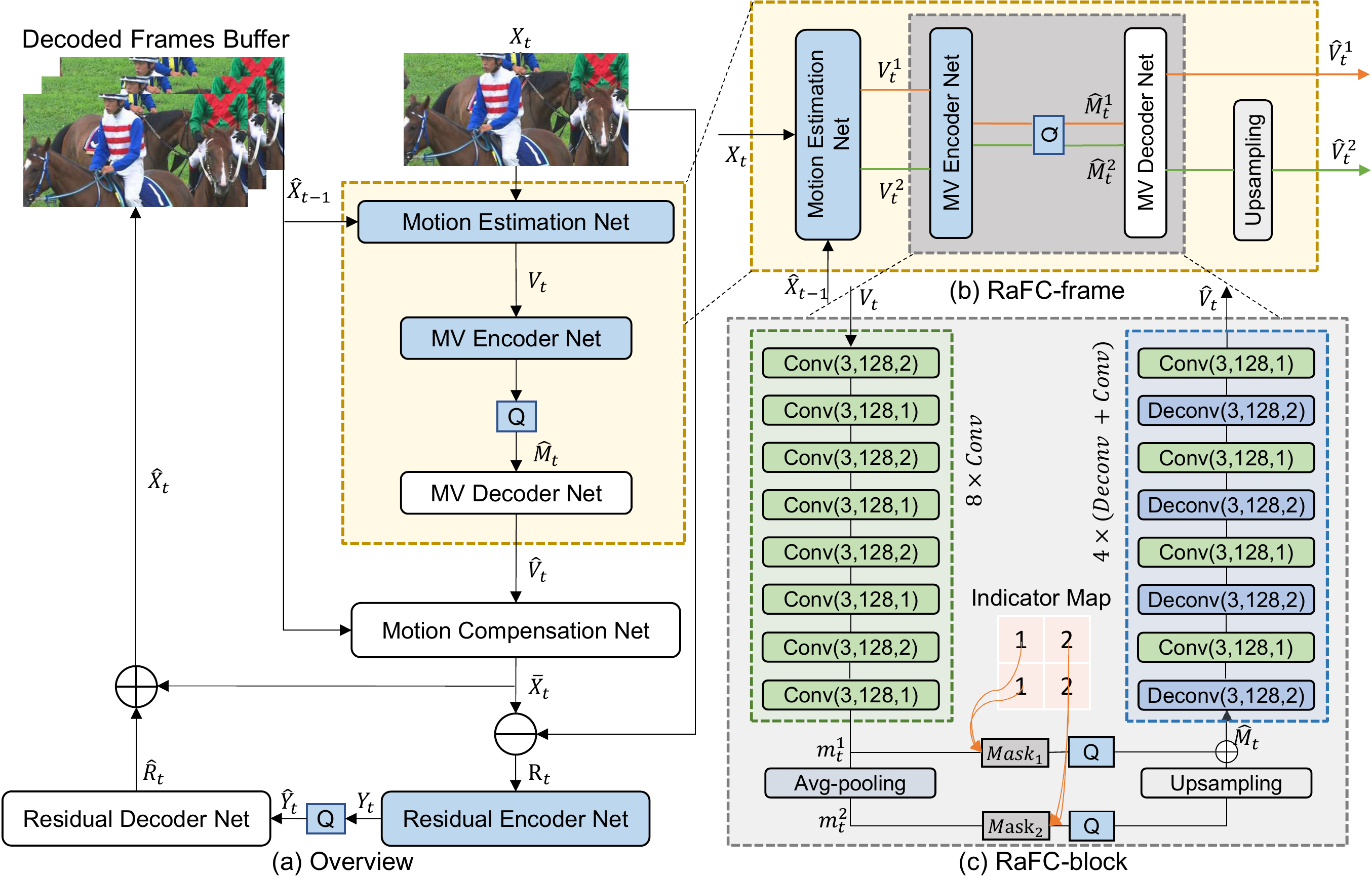}
% \end{center}
  \caption{Overview of our proposed framework and several basic modules used in our pipeline (a), the detailed motion coding modules in our frame-level scheme RaFC-frame (b) and our block-level scheme RaFC-block (c). In RaFC-frame (dashed yellow box), the ``Motion Estimation Net'' will generate two optical flow maps $V_t^{1}$ and $V_t^{2}$ with different resolutions and our method automatically select the optimal resolution (see the details in section~\ref{subsec:framel}(a)). In RaFC-block, the optical flow map $V_t$(i.e., $V_t^1$ or $V_t^2$) is transformed to multi-scale motion features $m_t^{1}$ and $m_t^{2}$, and we will select the most optimal resolution for each block by using the representations from either $m_t^{1}$ or $m_t^{2}$ to construct the reorganized motion feature $\hat{M}_t$, which will be used to obtain the reconstructed flow map $\hat{V}_t$ (see the details in section~\ref{subsec:blockl}(b)). In (c), Conv(3,128,2) represents the convolution operation with the kernel size of 3x3, the output channel of 128 and the stride of 2. Each convolution with the stride of 1 is followed by a Leaky ReLU layer. Two masks $Mask_1$ and $Mask_2$ are only used for ``Motion Feature Reorganization'' and are not used for ``Indicator Map Generation'' (see section~\ref{subsec:blockl}(b) for more details).}
\label{fig:overall}
\end{figure}

\section{Methodology}

\subsection{System Overview}
Figure~\ref{fig:overall}(a) provides an overview of the proposed video compression system. 
Inspired by the DVC~\cite{lu2019dvc} framework, we also use a hybrid coding scheme (e.g., motion coding and residual coding). 
The overall \textit{coding procedure} is summarized in the following steps.

\textit{\textbf{Motion coding.}}
We utilize our proposed RaFC method for motion coding.
RaFC consists of three modules, motion estimation net, the motion vector (MV) encoder net, and the MV decoder net. The motion estimation net estimates the optical flow $V_t$ between the input frame $X_t$ and the previous reconstructed frame $\hat{X}_{t-1}$ from the decoded frames buffer. Then, the MV encoder net encodes the optical flow maps as motion features/representations $M_t$, which is further quantized as $\hat{M}_t$ before entropy coding. Finally, the MV decoder net decodes the motion representation $\hat{M}_t$ so that the reconstructed flow map  $\hat{V}_t$ is obtained.

\textit{\textbf{Motion compensation.}}
Based on the reconstructed optical flow map $\hat{V}_t$ from the MV decoder and the reference frame $\hat{X}_{t-1}$,  a motion compensation network is employed to obtain the predicted frame $\bar{X}_t$. 

\textit{\textbf{Residual coding.}}
Denote the residual between the original frame $X_t$ and the predicted frame $\bar{X}_t$
by $R_t$.
Like in~\cite{lu2019dvc}, we adopt a residual encoder network to encode the residual as the latent representation $Y_t$ and then quantized as $\hat{Y}_t$ for entropy coding. Then the residual decoder network reconstructs the residual $\hat{R}_t$ from the latent representation $\hat{Y}_t$.

\textit{\textbf{Frame reconstruction.}}
With the predicted frame $\bar{X}_t$ from the motion compensation net and $\hat{R}_t$ obtained from the residual decoder net, the final reconstructed frame for $X_t$ can be obtained by $\hat{X}_t=\bar{X}_t+\hat{R}_t$, which is also sent to the decoded frames buffer and will be used as the reference frame for the next frame $X_{t+1}$.

\textit{\textbf{Quantization and Bit Estimation.}}
The generated latent representations (e.g., $\hat{Y}_t$) should be quantized before sending to the decoder side. 
To build an end-to-end optimized system, we follow the method in~\cite{balle2018variational} and add uniform noise to approximate quantization in the training stage.
Besides, we use the bitrate estimation network in~\cite{balle2018variational} to estimate the entropy coding bits. %To further improve the compression performance in motion coding, a better bitrate estimation network in~\cite{minnen2018joint} is applied in our framework.

In our proposed scheme, all the components in Figure~\ref{fig:overall}(a) are included in the encoder side, and only the MV decoder net, motion compensation net and residual decoder net are used in the decoder side.

% \{-5mm}
\subsection{Problem Formulation}
We use 
$X = \{X_1, X_2,...,X_{t-1}, X_t,...\}$ to denote the input video sequence to be compressed, where $X_t \in \mathbb{R}^{W\times H \times C}$ represents the frame at time step $t$. $W$, $H$, $C$ represent the width, the height and the number of channels (i.e., $C=3$ for RGB videos). 
Given the input video sequences, the video encoder will generate the corresponding bitstreams, while the decoder reconstructs the video sequences by using the received bitstreams.
To achieve highly efficient compression, the whole video compression system needs to generate high quality reconstructed frames at any given bitrate budget.
Therefore, the objective of the learning based video compression system is formulated as follows,

% \{-2mm}
\begin{equation}
\label{eq:rd}
    RD = R + \lambda D =  (\mathbb{H}(\hat{M}_t) + \mathbb{H}(\hat{Y}_t)) + \lambda d(X_t, \hat{X}_t),
\end{equation}
The term $R$ in Eq. (\ref{eq:rd}) denotes the number of bits used to encode the frame.
$R$  is calculated by adding up the number of bits $\mathbb{H}(\hat{M}_t)$  for encoding the flow information and the number of bits $\mathbb{H}(\hat{Y}_t)$ for encoding the residual information.
$D = d(X_t, \hat{X}_t)$ denotes the distortion between the input frame and the reconstructed frame,
where $d(\cdot)$ represents the metric (mean square error or  MS-SSIM~\cite{wang2003multiscale}) for measuring the difference between two images.

In the traditional video compression system, the rate-distortion optimization (RDO) technique is widely used to select the optimal mode for each coding block. The RDO procedure is formulated as follows,
\begin{equation}
    \mathcal{M} = \argmin_{i \in \mathcal{C}} RD_{i}
\label{eq:rdo_t}
\end{equation}
where $RD_{i}$ represents the RD value of the $i^{th}$ mode, and
$\mathcal{C}$ represents the candidate modes.
The RDO procedure will select the optimal mode $\mathcal{M}$ with the minimum rate-distortion (RD) value to achieve highly efficient video coding.

However, this basic technique is not exploited in the state-of-the-art learning based video compression systems. In this work, we propose the RaFC framework to effectively compress motion information by using multi-resolution representations for the flow maps and motion features.
%in learned video codec. 
The key idea in our method is to use the RDO technique to select the optimal resolution of optical flow maps or motion features at each block for the current frame.
% As far as we know, we are the first learning based video compression method that uses the RDO technique to achieve high efficiency video coding.% in the inference phase.

% \{-5mm}

\subsection{Resolution-adaptive Flow Coding (RaFC)}
In this section, we introduce our RaFC scheme for motion compression and present how to select the optimal flow map or motion features by using the RDO technique based on the RD criterion.

% \subsubsection{Frame-level scheme RaFC-frame}
\textbf{(a) Frame-level scheme RaFC-frame}

\label{subsec:framel}

As shown in Figure~\ref{fig:overall}(b), given the input frame $X_t$ and its corresponding reference frame $\hat{X}_{t-1}$ from the decoded frames buffer, we utilize the motion estimation network to generate the multi-scale flow maps.
Taking advantage of the existing pyramid architecture in Spynet~\cite{ranjan2017optical} in our work, we generate two flow maps $V^{1}_t$ and $V^{2}_t$ with the resolutions of $W \times H$ and $\frac{W}{2} \times \frac{H}{2}$, respectively.
While more resolutions can be readily used in our RaFC-frame method, we observe that our RaFC-frame scheme based on two-scale optical flow maps has already been able to achieve promising results.

In our proposed frame-level scheme RaFC-frame, the goal is to select the optimal resolution from the multi-scale optical flow maps for the current frame in order to handle complex or simple motion patterns globally.
According to the RDO formulation in Eq.~\eqref{eq:rdo_t}, we need to calculate the RD values for the two optical flow maps $V^{1}_t$ and $V^{2}_t$ respectively. The details are provided below.

\textbf{Calculating the rate-distortion(RD) value.}
We take the optical flow map $V^{2}_t$ as an example to introduce how to calculate the RD value. First,  as shown in Figure~\ref{fig:overall}(b), based on the MV encoder and the MV decoder, we can obtain the reconstructed optical flow map and the corresponding quantized representation $\hat{M}^{2}_t$. While the resolution of the reconstructed flow map is only $\frac{W}{2} \times \frac{H}{2}$, there is an additional upsampling operation before obtaining $\hat{V}_t^2$, so the resolution of $\hat{V}_t^2$ is also $W \times H$. 
After going through the subsequent coding procedure, such as the motion compensation unit, the residual encoder unit and the residual decoder unit (see Section 3.1 for more details), we arrive at the reconstructed frame $\hat{X}_t^2$ and also obtain the corresponding bitstreams from $\hat{M}^{2}_t$ and $\hat{Y}_t^2$, for motion information and residual information, respectively.
Therefore, based on Eq.~\eqref{eq:rd}, we can calculate the RD value for the flow map $V^{2}_t$. We can similarly calculate the RD value for the flow map $V_t^{1}$. Finally, we select the optimal flow map with the minimum RD value.

After selecting the optimal flow map of the current frame by using the RDO technique in Eq.~\eqref{eq:rdo_t}, we can update the network parameters by using the loss function defined in Eq.~\eqref{eq:rd}, where $\hat{M}_t$, $\hat{Y}_t$ and $\hat{X}_t$ are obtained based on the selected flow map (i.e., $V^1_t$ or $V^2_t$).

\begin{figure}[t]
\begin{center}
  \includegraphics[width=0.6\linewidth]{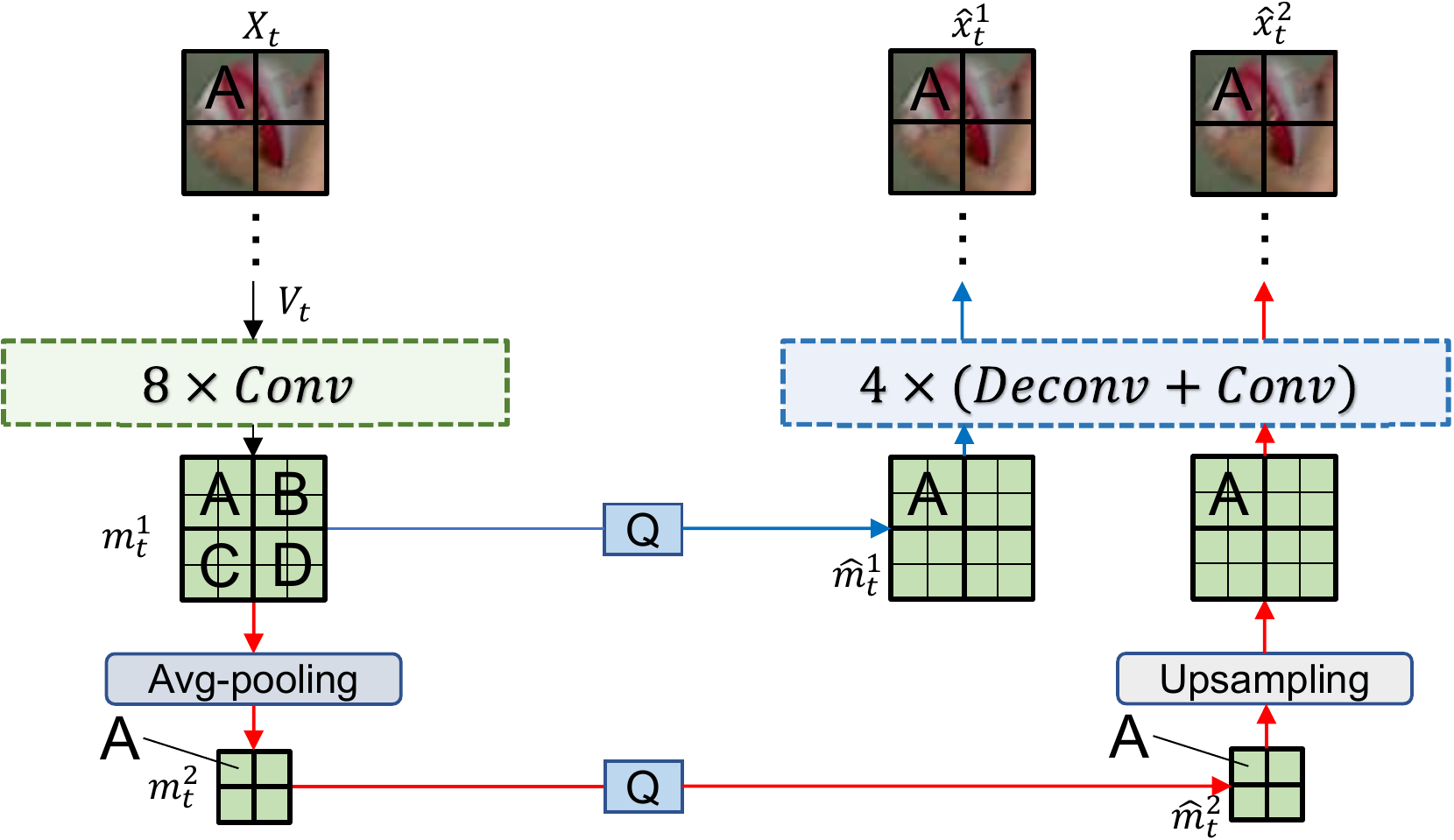}
\end{center}
  \caption{Generation of the indicator map. The network structures of $8\times Conv$ and $4\times(Deconv+Conv)$ are provided in Figure~\ref{fig:overall}(c). For better illustration, one channel is shown as an example. }
\label{fig:indicator}
\end{figure} 

% \subsubsection{Block-level scheme RaFC-block}
\textbf{(b) Block-level scheme RaFC-block}

Previous learning based video compression systems only use motion features with fix resolution to represent optical flow information.
In H.264 and H.265, different block sizes are used for motion estimation.
To this end, it is necessary to design an efficient multi-scale motion features in order to handle different types of motion patterns.%fast and slow movement of objects.

As shown in Figure~\ref{fig:overall}(c), given the optical flow map $V_t$ from one resolution (i.e. $V_t$ can be $V_t^{1}$ or $V_t^{2}$ from Section ~\ref{subsec:framel}(a)), we firstly feed the optical flow map $V_t$ to
generate the multi-scale motion features $m^1_t$ and $m^2_t$. 
Here we just use two-resolution motion features as an example, and our approach can be readily used for more resolutions (we use three-resolution motion features in our experiments).
Then, the proposed  RaFC-block method will select the optimal resolution of the motion features for each block in the reconstructed frame based on the RDO technique. Specifically, we proposed a two-step procedure, which is summarized as follows.

\textbf{Indicator Map Generation.}
In Figure~\ref{fig:indicator}, we take an input image with the resolution of $64\times 64$ as an example to introduce how to generate the indicator map with the size of $2\times2$. After four pairs of convolution layers with the strides of 1 and 2, we can obtain the motion feature $m_t^1$ with the resolution of $4\times 4$. We divide $m_t^1$ as 4 blocks A, B, C and D, and each block represents a $2\times2$ region. Based on $m_t^1$, we further obtain $m_t^2$ with the resolution of $2\times2$ after going through another average pooling layer. Then for each block (A, B, C,  or D), we need to decide whether we should choose the $2\times2$ representation from $m^{1}_t$ or the $1\times1$ representation from $m^{2}_t$. The details are provided below.

After quantizing $m_t^1$ to obtain $\hat{m}_t^{1}$, we will go through four pairs of deconvolution and convolution layers and the rest coding procedure (e.g. the motion compensation unit, the residual encoder unit and the residual decoder unit), we can obtain the final reconstructed image $\hat{x}_t^1$ with the resolution of $64\times64$ from $\hat{m}_t^1$. We also quantize $m_t^2$ as $\hat{m}_t^2$, and go through an additional upsampling layer to reach the same size with $\hat{m}_t^1$. Then after four pairs of deconvolution and convolution layers and the rest coding procedure, we can also obtain $\hat{x}_t^2$ with the resolution of $64\times64$. We then similarly divide $\hat{x}_t^1$ and $\hat{x}_t^2$ as four blocks A, B, C, and D. For each block in both $\hat{x}_t^1$ and $\hat{x}_t^2$, we can calculate the RD value by using Eq.~\eqref{eq:rd}, where the bit rates are calculated by using the corresponding motion features and the residual image at one specific block, and the distortion D is also calculated for this specific block. By choosing the smaller RD value, we can determine which representation of motion feature (i.e., the $2\times2$ representation from $m^{1}_t$ or the $1\times1$ representation from $m^{2}_t$) will be used at each block.

In this way, we can obtain the indicator map which represents the optimal resolution choice at each block. While more advanced approaches can be used to decide the indicator map, it is worth mentioning that the aforementioned solution is efficient and achieves promising results (see our results in Section 4).

\begin{figure}[t]
\begin{center}
  \includegraphics[width=0.6\linewidth]{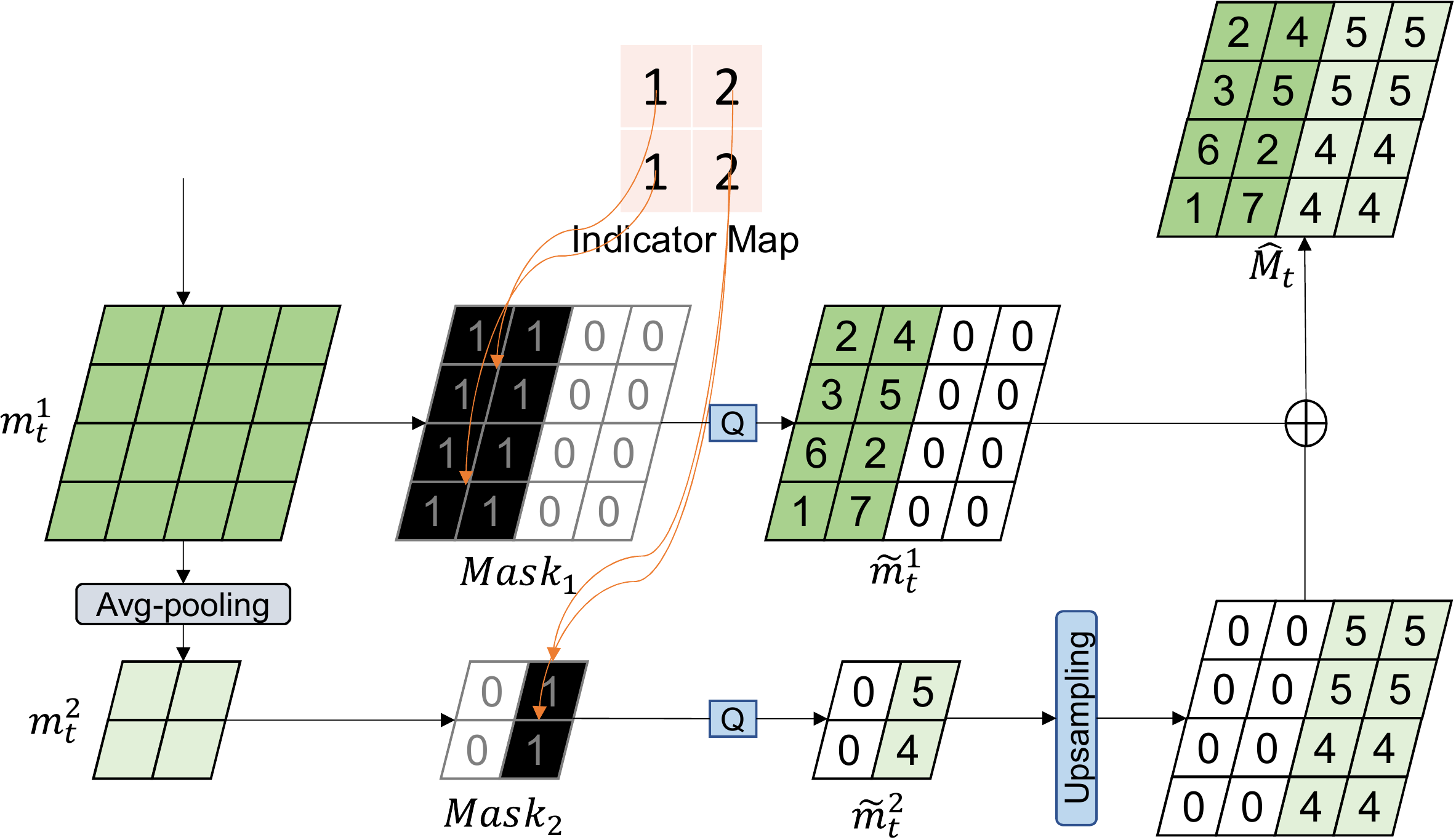}
\end{center}
  \caption{Motion feature reorganization with the indicator map. For better illustration, one channel is shown as an example.}
\label{fig:combine}
\end{figure} 

\textbf{Motion Feature Reorganization.}
\label{subsec:blockl}
In our approach, we need to reorganize the motion representation based on the indicator map. 
As shown in Figure~\ref{fig:combine}, given the indicator map and the quantized features, we first obtain the masked and quantized multi-scale motion features $\Tilde{m}^1_t$ and $\Tilde{m}^2_t$. 
The corresponding locations without features, which are also masked at the encoder side, are filled with zeros.
Then from bottom to top, $\Tilde{m}^2_t$ is first upsampled to the same size of $\Tilde{m}^1_t$, which is then added to $\Tilde{m}^1_t$. 
In this way, we can obtain the reorganized motion feature $\hat{M}_t$, which exploits the multi-scale motion representations for better motion compression.

After motion feature reorginzation, we can easily obtain the quantized residual information $\hat{Y}_t$ and the reconstructed frame $\hat{X}_t$ by following the hybrid coding scheme in Figure~\ref{fig:overall}(a), which includes the motion compensation unit, the residual encoder unit and the residual decoder unit. Then the loss function defined in Eq.~\eqref{eq:rd} will be minimized to update the network parameters.

% \subsubsection{Our overall RaFC framework by combining both schemes}
\textbf{(c) Our overall RaFC framework by combining both schemes}

The frame-level scheme RaFC-frame selects the optimal resolution of optical flow maps, which is the input of the MV encoder, while the block-level scheme RaFC-block selects the optimal resolution for motion features at each block, which is the output of the MV encoder.   Therefore, these two techniques are complementary to each other and can be readily combined.
 
Specifically, we embed the block-level method RaFC-block into the frame-level method RaFC-frame.  For the first input flow map $V_t^1$, we use the RaFC-block method to decide the optimal indicator map based on the RD criterion at the block level, and then output $\hat{V}_t^1$ based on the reorganized motion feature. After going through the subsequent coding process including the motion compensation unit, the residual encoder unit and the residual decoder unit, we finally obtain the reconstructed frame $\hat{X}_t^1$. Based on the distortion between $\hat{X}_t^1$ and $X_t$, and the numbers of bits used for encoding both the reorganized motion feature and residual information, we can calculate the RD value. For the second input flow map $V_t^2$, we perform the same process and calculate the RD value. Finally, we choose the optimal mode with the minimum RD value for encoding motion information of the current frame. Here, the optimal mode includes the selected optical flow map and the corresponding selected resolution of motion features at each block for this selected flow map.  
 
After selecting the optimal mode for encoding the motion information of the current frame, we update all the parameters in our network by minimizing the objective function in Eq.~(\ref{eq:rd}), where the distortion and the numbers of bits used to encode the motion features and the residual information are obtained for the selected mode. 

% \{-5mm}
\section{Experiment}

\subsection{Experimental Setup}

\textbf{Datasets.}
We use the Vimeo-90k dataset~\cite{xue2019video} to train our framework and each clip in this dataset consists of 7 frames with the resolution of $448 \times 256$.

For performance evaluation, we use four datasets: HEVC Class E~\cite{sullivan2012overview} , UVG~\cite{UVGdataset}, MCL-JCV~\cite{wang2016mcl} and VTL~\cite{VTLdataset}. 
% The HEVC Standard Test Sequences has been widely used for evaluating the traditional video compression methods and we use HEVC Class E in our experiments. 
The HEVC Standard Test Sequences have been widely used for evaluating the traditional video compression methods, in which the HEVC class E dataset contains three videos with the resolution of $1280 \times 720$. 
The UVG dataset~\cite{UVGdataset} has seven videos with the resolution of $1920 \times 1080$. 
The MCL-JCV dataset~\cite{wang2016mcl} has been widely used for video quality evaluation, which has 30 videos with the resolution of $1920 \times 1080$.
%
% The VTL dataset~\cite{VTLdataset} also has the sequences with different resolutions. 
For the VTL dataset~\cite{VTLdataset}, we follow the experimental setting in~\cite{abdelaziz2019neural} and use the first 300 frames in each video clip for performance evaluation.

\textbf{Evaluation Metric.}
We use PSNR and MS-SSIM~\cite{wang2003multiscale} to measure the distortion between the reconstructed and ground-truth frames.
PSNR is the most widely used metric for measuring compression distortion,
while MS-SSIM has been adopted in many recent works to evaluate the subjective visual quality.
%
% We use bit per pixel (Bpp) to denote the average number of bits for one pixel in each image.
We use bit per pixel (Bpp) to denote the bitrate cost in the compression procedure.

\begin{figure}
  \centering
  \begin{minipage}[c]{0.5\textwidth}
    \centering
    \includegraphics[height=1.75in]{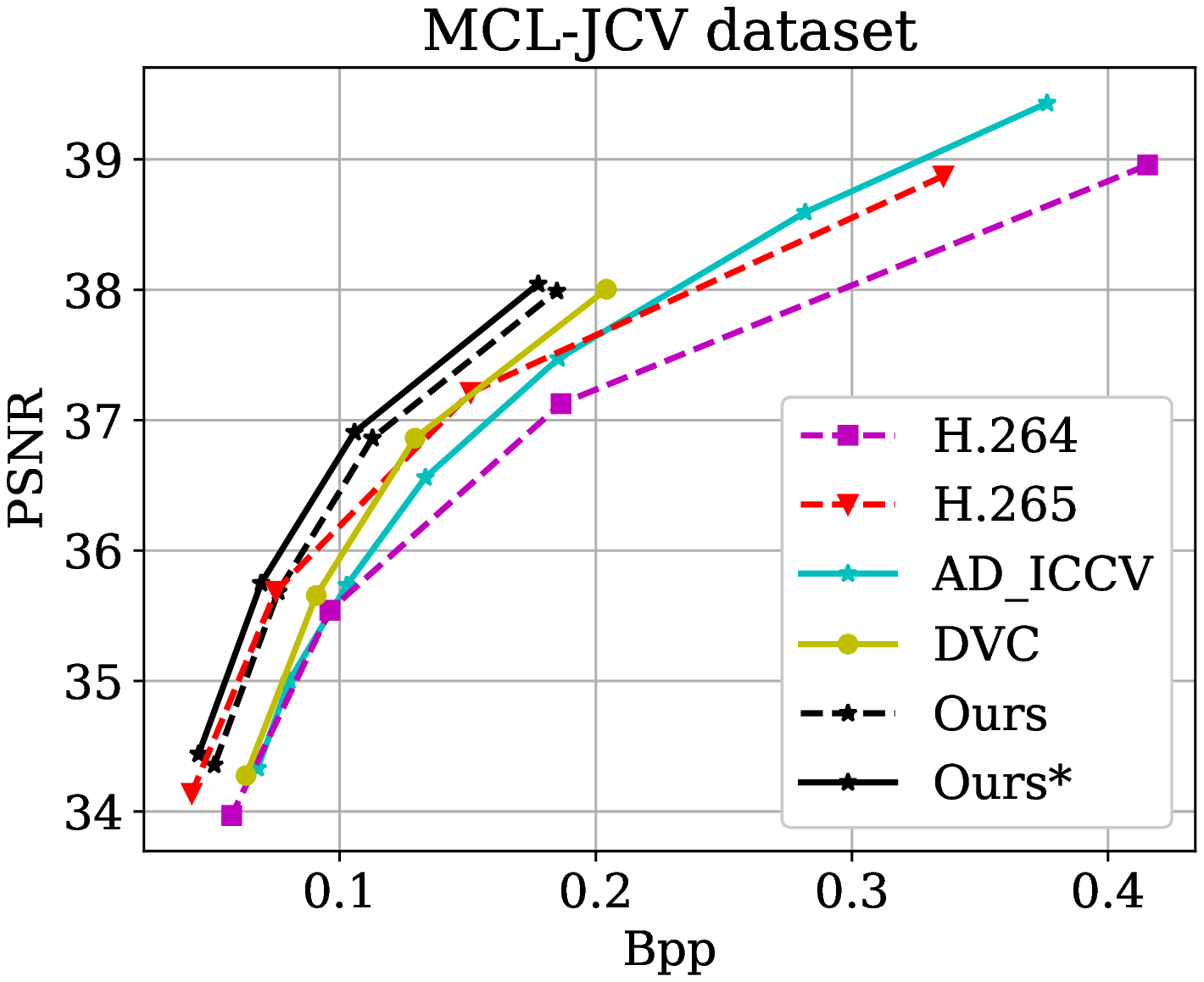}
  \end{minipage}%
  \begin{minipage}[c]{0.5\textwidth}
  \centering
    \includegraphics[height=1.75in]{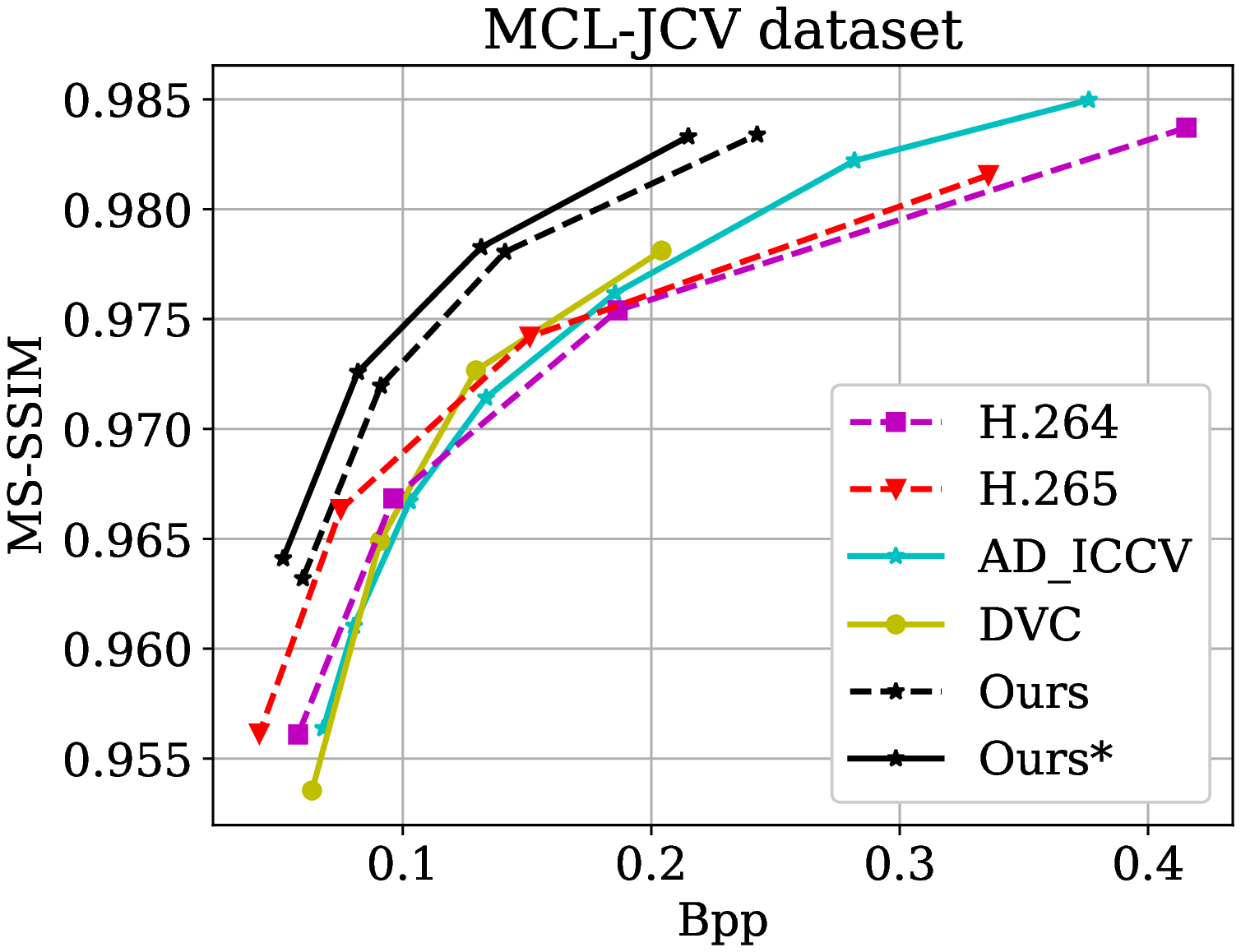}
  \end{minipage}
  \begin{minipage}[c]{0.5\textwidth}
  \centering
    \includegraphics[height=1.75in]{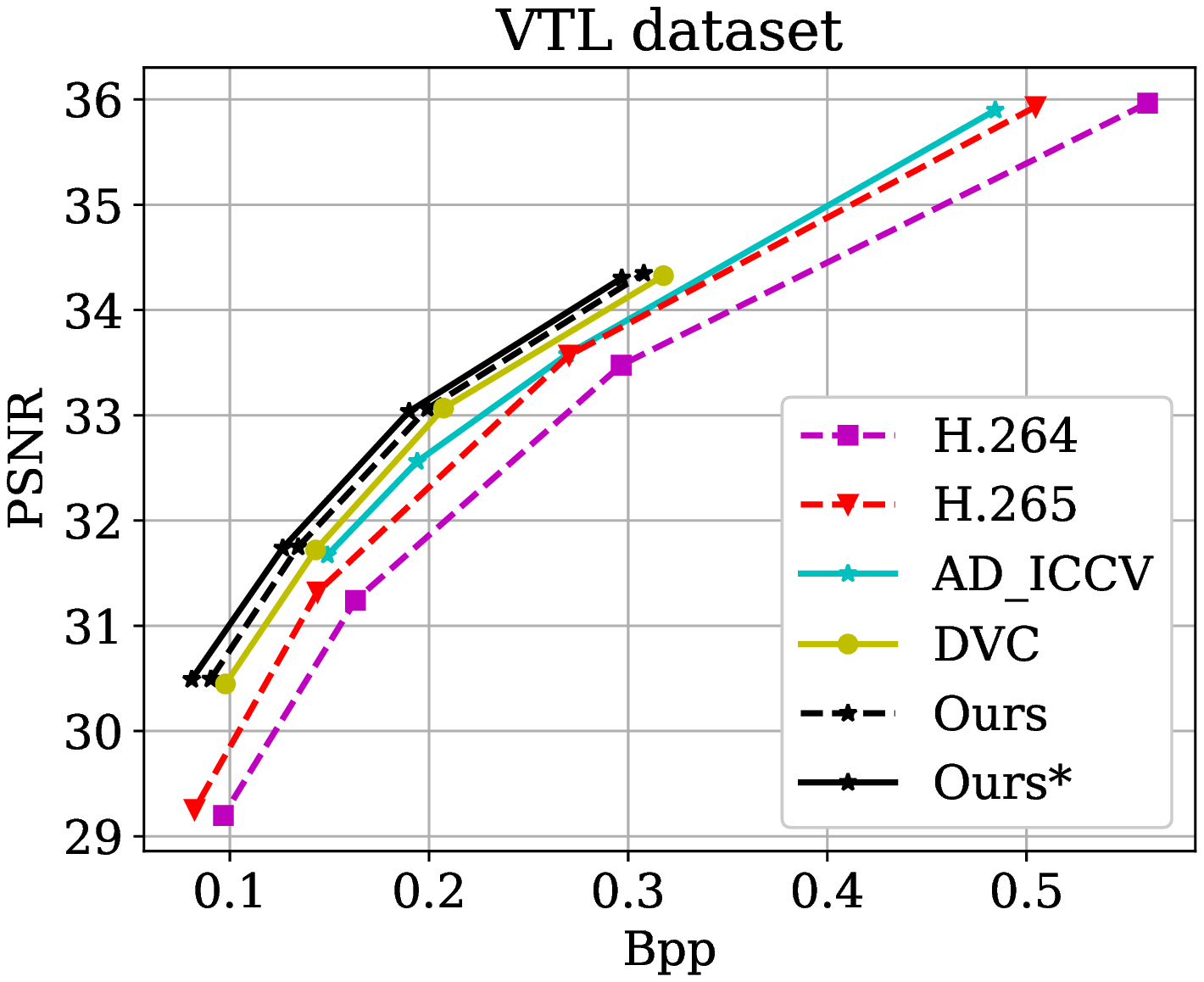}
  \end{minipage}%
  \begin{minipage}[c]{0.5\textwidth}
  \centering
    \includegraphics[height=1.75in]{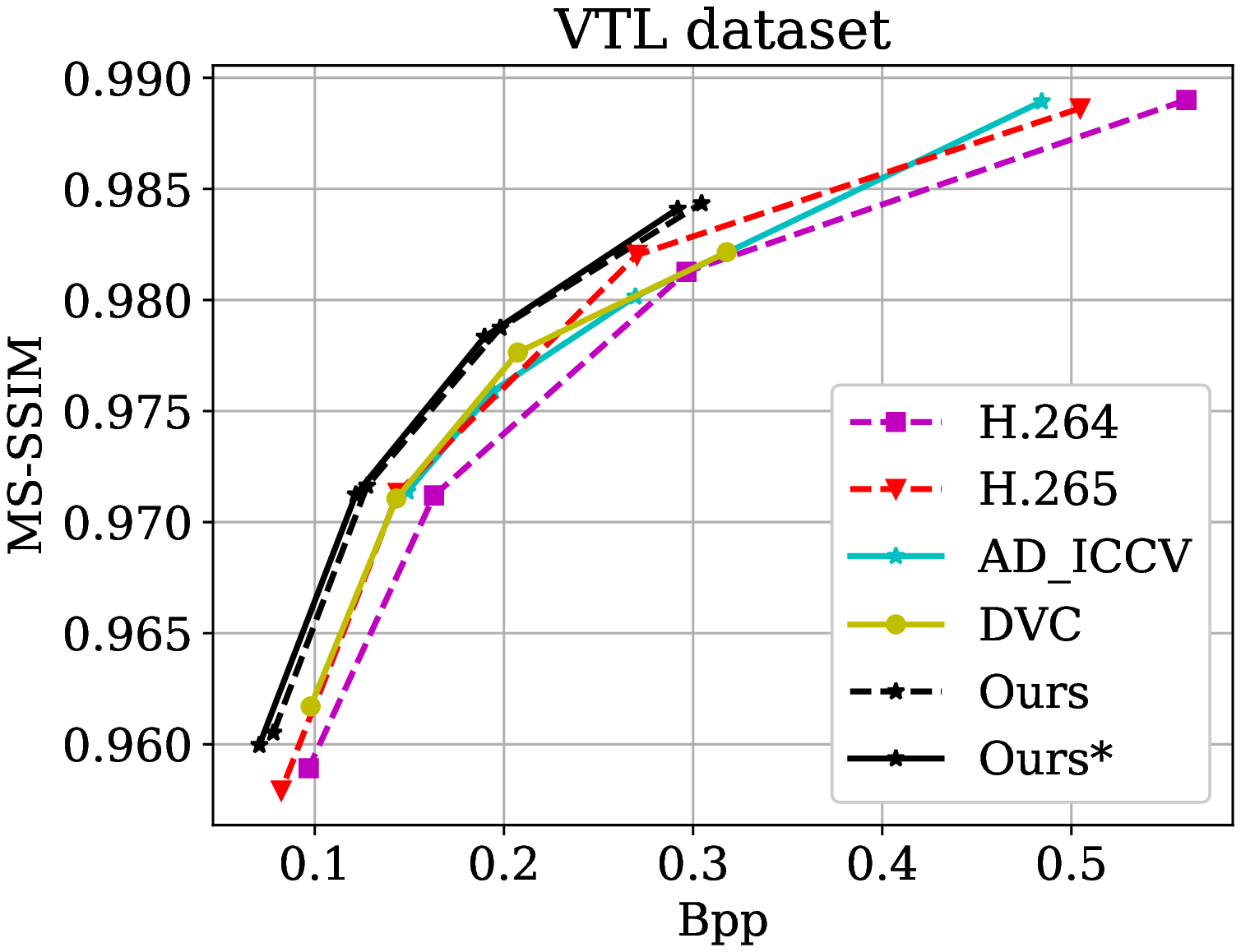}
  \end{minipage}
  \centering
  \begin{minipage}[c]{0.5\textwidth}
    \centering
    \includegraphics[height=1.75in]{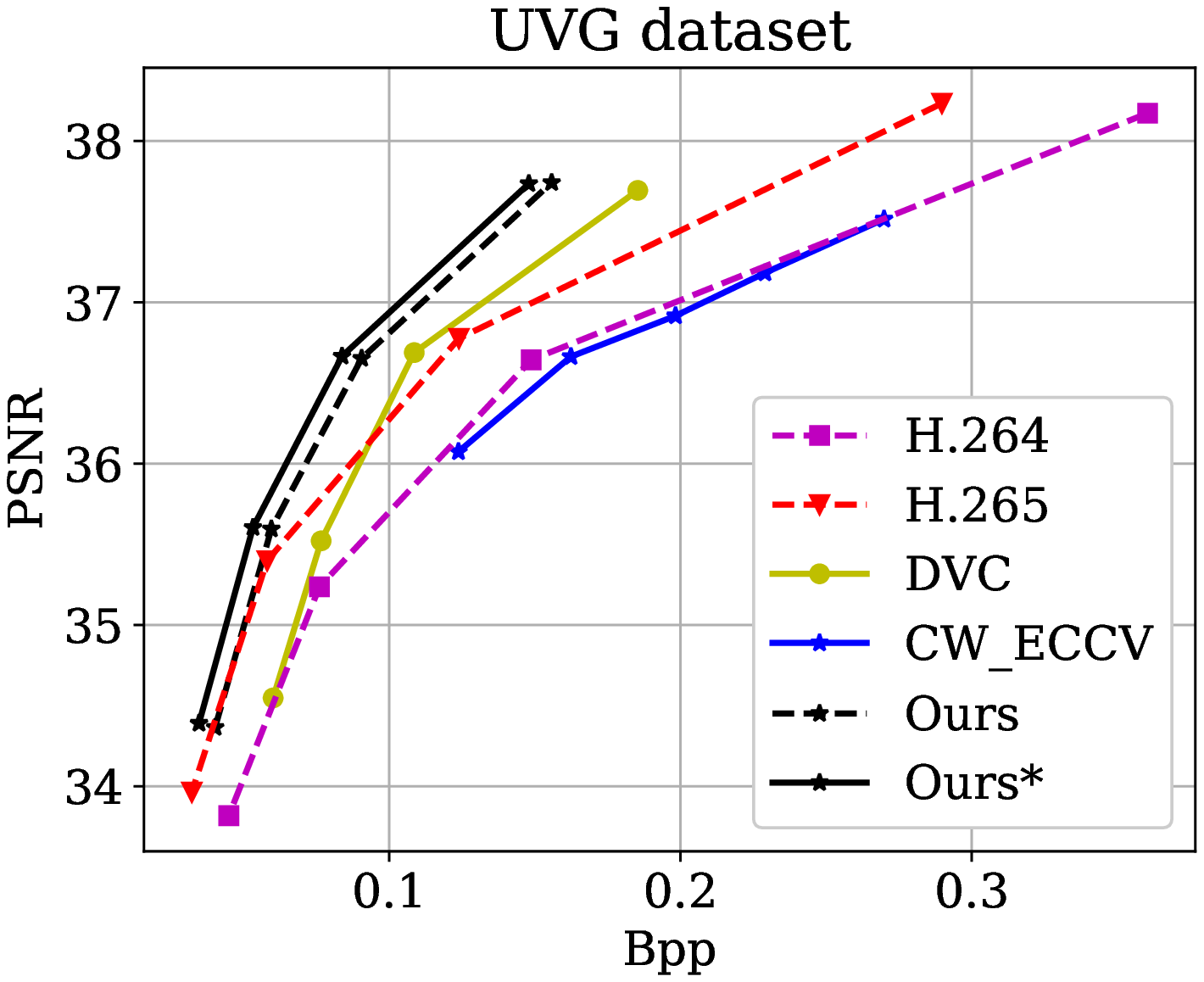}
  \end{minipage}%
  \centering
  \begin{minipage}[c]{0.5\textwidth}
    \centering
    \includegraphics[height=1.75in]{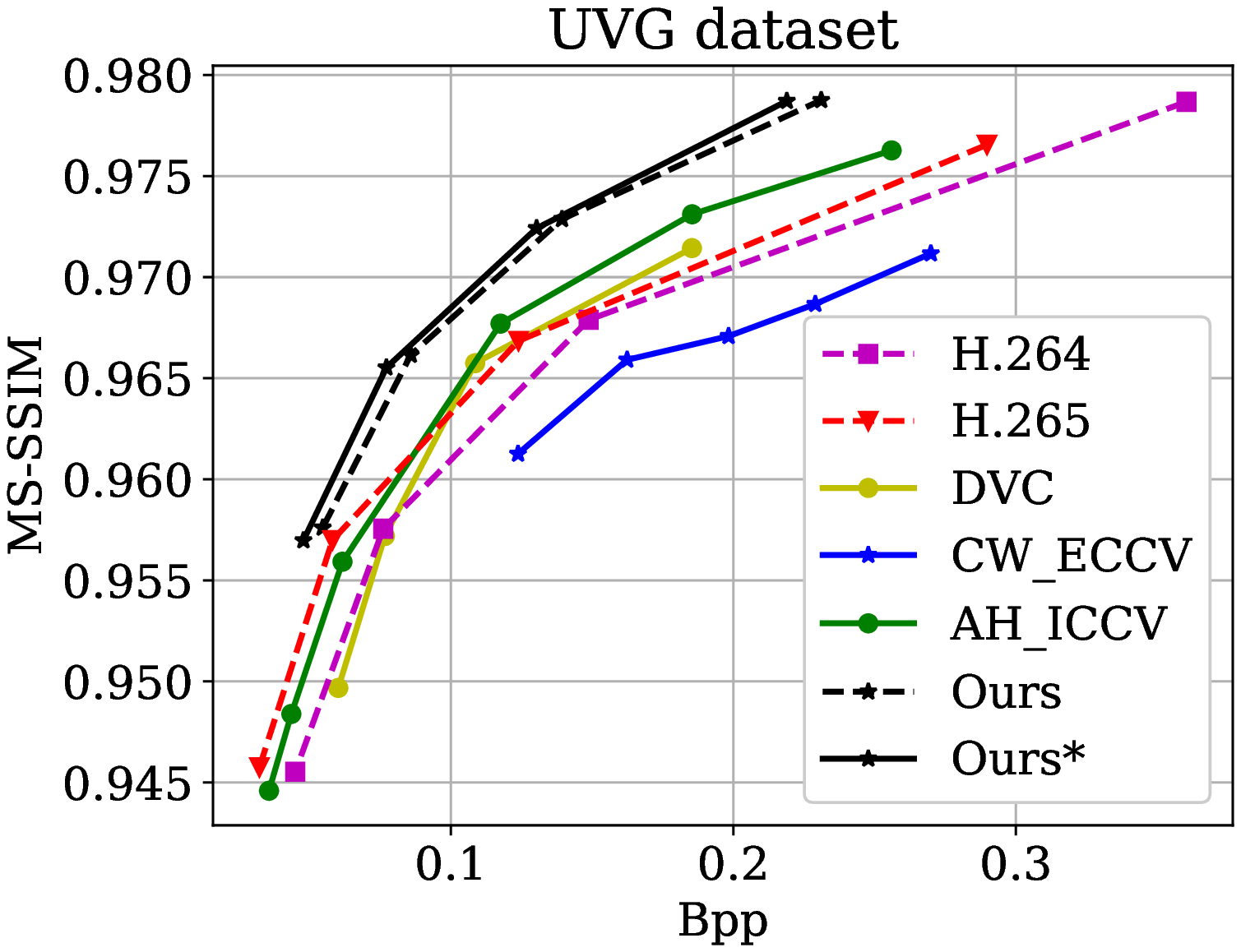}
  \end{minipage}
  \begin{minipage}[c]{0.5\textwidth}
  \centering
    \includegraphics[height=1.75in]{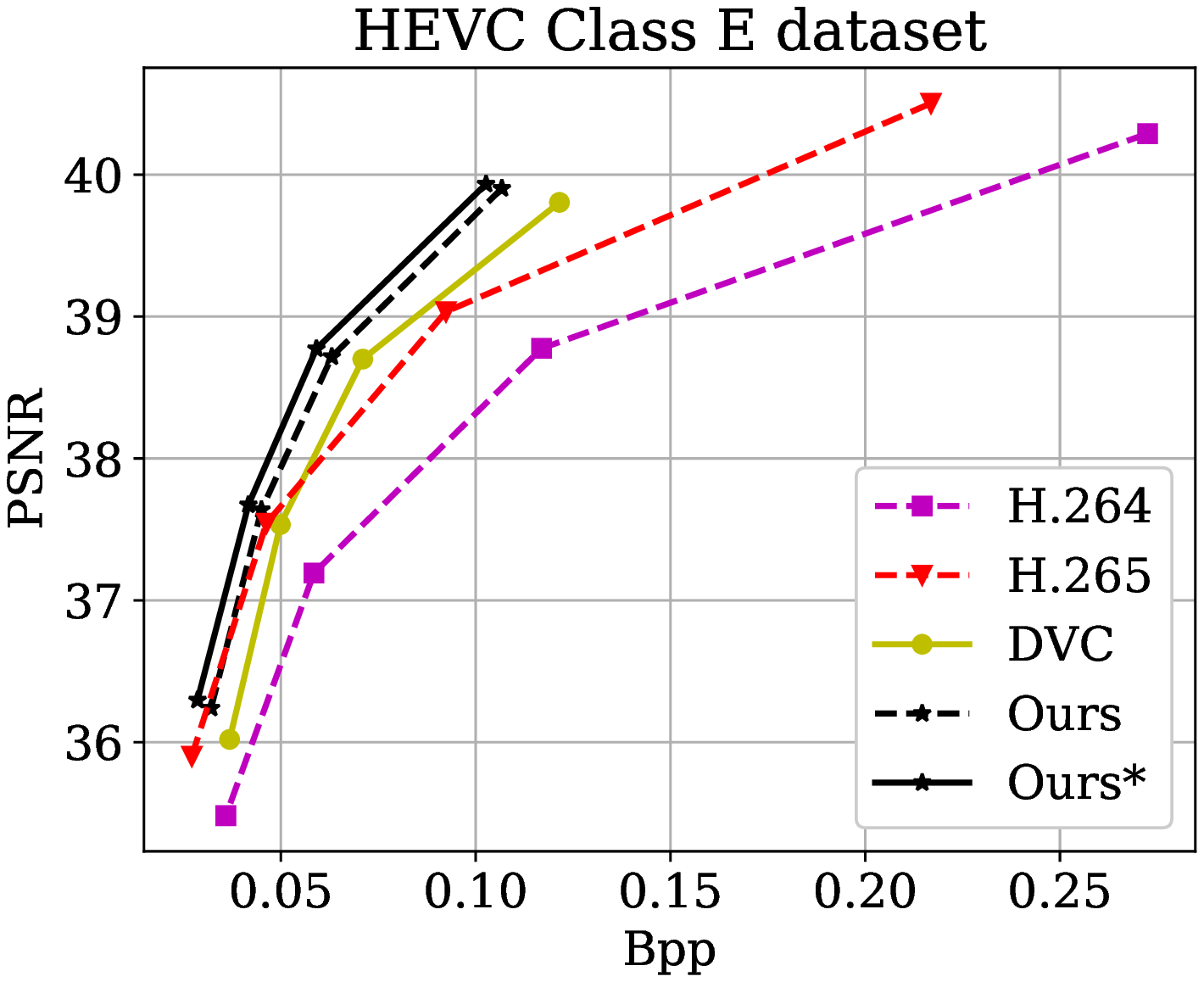}
  \end{minipage}%
  \begin{minipage}[c]{0.5\textwidth}
  \centering
    \includegraphics[height=1.75in]{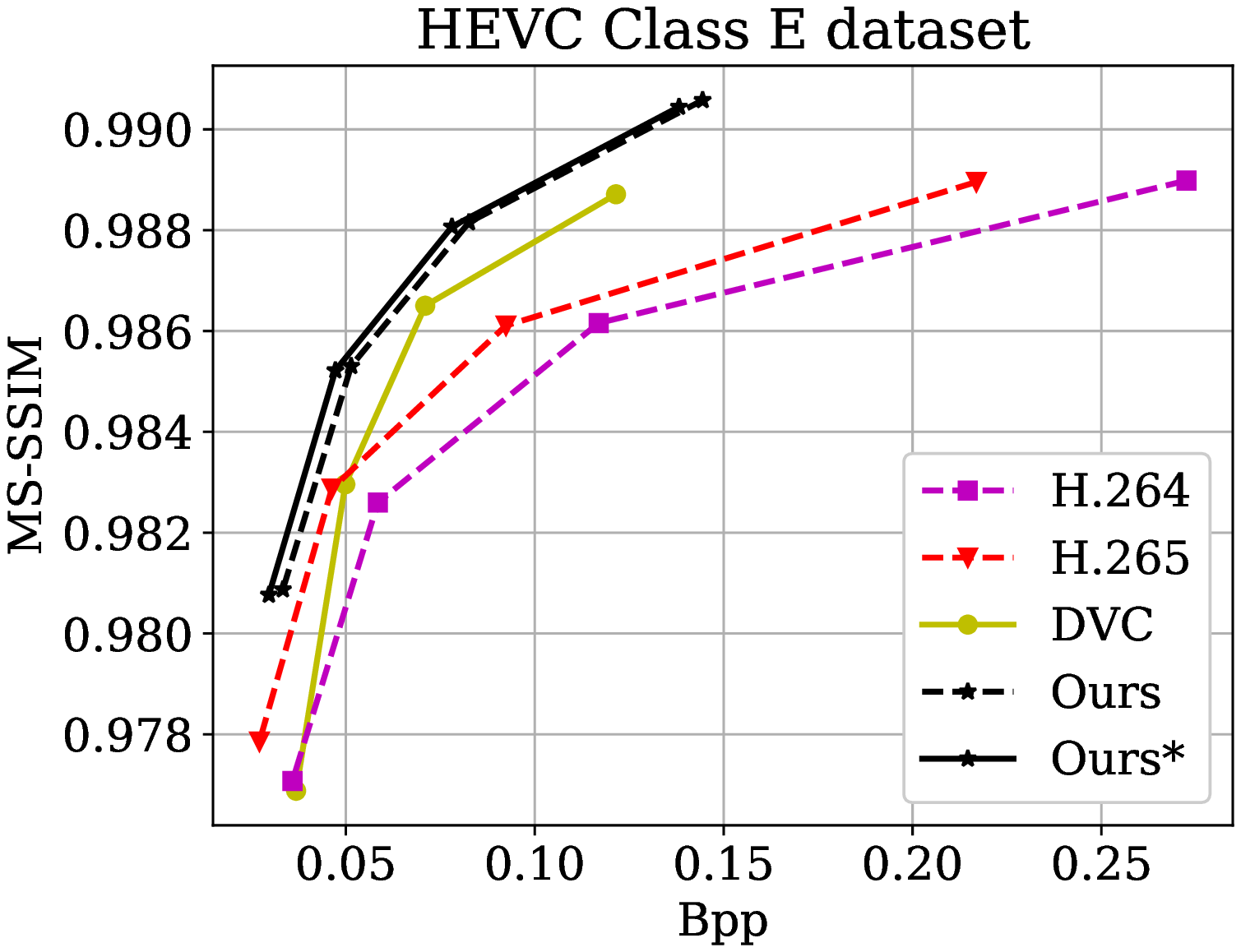}
  \end{minipage}
    \caption{Experimental results on the MCL-JCV, VTL, UVG and HEVC Class E datasets.}
  \label{fig:result}
\end{figure}

\textbf{Implementation Details.}
We train our model in two stages.
At the first stage, we set $\lambda$ as 2048, and train our model based on mean square error for 2,000,000 steps to obtain a pre-trained model at high bitrate.
At the second stage, for different $\lambda$ values ($\lambda =  256, 512, 1024$ and $2048$), we fine-tune the pretrained model for another 500,000 iterations. 
% \textcolor{red}{
% Additionally, we use the RD values, which are used in the selection process of our methods RaFC-frame and RaFC-block as part of our training loss for the first 250,000 steps.
%
% Specifically, we use mean square error (MSE) as our distortion function.% for the first 2,500,000 steps.}
To achieve better MS-SSIM performance, we additionally fine-tune the models from the second stage for about 80,000 steps by using the MS-SSIM criterion as the distortion term when calculating the RD values. %In addition, multiple consecutive frames (three frames are employed in our experiment) are used as a small video clip for training and only the loss from the last time step is used for back-propagation during the training process.

Our framework is implemented based on Pytorch with CUDA support.
% All the experiments are conducted on  NVIDIA 2080TI GPU with 11GB memory. %
In the training phase, we set the batch size as 4.
We use the Adam optimizer~\cite{kingma2014adam} with the learning rate of 1e-4 for the first 1,800,000 steps and  1e-5 for the remaining steps.
It takes about 6 days to train the proposed model.

In our experiments,  motion features ($\hat{m}_t^1$, $\hat{m}_t^2$ and $\hat{m}_t^3$) with three different resolutions are used in our RaFC-block module (note $\hat{m}_t^3$ can be similarly obtained from $\hat{m}_t^2$ as shown in Fig 2).
It is noted that one pixel in $\hat{m}_t^1$, $\hat{m}_t^2$ and $\hat{m}_t^3$ correspond to one block with the resolution of $16\times16$, $32\times32$ and $64\times64$ in the original optical flow map, respectively.

% \{-5mm}
\subsection{Experimental Results}

% \{-1mm}
The experimental results on different datasets are provided in Figure~\ref{fig:result}.
% The proposed method using the same entropy models (i.e., in~\cite{balle2018variational} ) as baseline method DVC~\cite{lu2019dvc} is denoted as ``Ours", while "Ours*" represents the proposed framework with better entropy model~\cite{minnen2018joint}.
% To further improve the compression performance in motion coding, a better bitrate estimation network in~\cite{minnen2018joint} is applied in our framework and the results are denoted as "Ours*".
% The proposed method using the same entropy models as baseline method DVC~\cite{lu2019dvc} is denoted as ``Ours", while the ``Ours*" represents the proposed framework using better entropy model~\cite{minnen2018joint} in the motion compression.
In DVC~\cite{lu2019dvc}, the hyperprior entropy model~\cite{balle2018variational} is used to compress the flow maps. However, other advanced methods like the auto-regressive entropy model~\cite{minnen2018joint} can be readily used to compress the flow maps. To this end, we report two results for our RaFC framework, which are denoted as ``Ours” and ``Ours*”. In ``Ours”, the hyperprior entropy model~\cite{balle2018variational} is incorporated in our RaFC framework in order to fairly compare our RaFC framework with DVC. In ``Ours*”, the auto-regressive entropy model~\cite{minnen2018joint} is incorporated in our RaFC framework to further improve the video compression performance. 
% The results for HEVC Class B, C and D are provided in the supplementary material.
We use the traditional compression methods H.264~\cite{wiegand2003overview}, H.265~\cite{sullivan2012overview} and the state-of-the-art learning-based compression methods, including DVC~\cite{lu2019dvc}, AD\_ICCV~\cite{abdelaziz2019neural}, AH\_ICCV~\cite{habibian2019video} and CW\_ECCV~\cite{wu2018video} for performance comparison.
It is noted that CW\_ECCV~\cite{wu2018video} and AD\_ICCV~\cite{abdelaziz2019neural} are B-frame based compression methods, while the others are P-frame based compression methods.
For H.264 and H.265, we follow the setting in DVC~\cite{lu2019dvc} and use FFmpeg with the \textit{default} mode. 
%
%The GOP size for the HEVC dataset is set to 10 and it is set as 12 for the VTL dataset, UVG dataset and MCL-JCV dataset. 
% The details of the H.264/H.265 settings in our work are provided in the supplementary material. 
We use the image compression method~\cite{balle2018variational} to reconstruct the I-frame.%Intra-frame. 
% The MS-SSIM results of our model reported in Fig.\ref{fig:result} are based on 

% For all MS-SSIM results, the fine-tuning scheme is adopted by using the MS-SSIM criterion to measure the distortion when calculating the RD value.

As shown in Figure~\ref{fig:result}, our method using the hyperprior entropy model (i.e., ``Ours") outperforms the baseline method DVC on all datasets, which demonstrates it is beneficial to use our newly proposed framework RaFC to compress the optical flow maps.
In other words, it is necessary to choose the optimal resolutions for the optical flow maps and the corresponding motion features in video compression.
When compared with our method using the hyperprior entropy model (i.e., ``Ours"), our method using the auto-regressive entropy model (i.e., ``Ours*") further improves the results, which demonstrates the effectiveness of the auto-regressive entropy model for flow compression.
% by a large margin on the UVG dataset and the HEVC Class E dataset. 
%
Our method using the auto-regressive entropy model~\cite{balle2018variational} achieves the best results on all datasets. Specifically, our method~(i.e.,``Ours*") has about 0.5dB gain over DVC at 0.1bpp on the UVG dataset.
On the MCL-JCV dataset, our approach~(i.e.,``Ours*") outperforms the interpolation based video compression method AD\_ICCV in terms of both PSNR and MS-SSIM. In addition, it also achieves about 0.4dB improvement at 0.2bpp over AD\_ICCV on the VTL dataset in terms of PSNR. 
Although our method is designed for P-frame compression, we can still achieve better compression performance than the B-frame compression methods AD\_ICCV and CW\_ECCV, which demonstrates the effectiveness of our approach.

\begin{figure}[!t]
\centering
\subfigure[Ablation study on the UVG dataset. DVC is adopted as our baseline method.]{\includegraphics[height=1.40in]{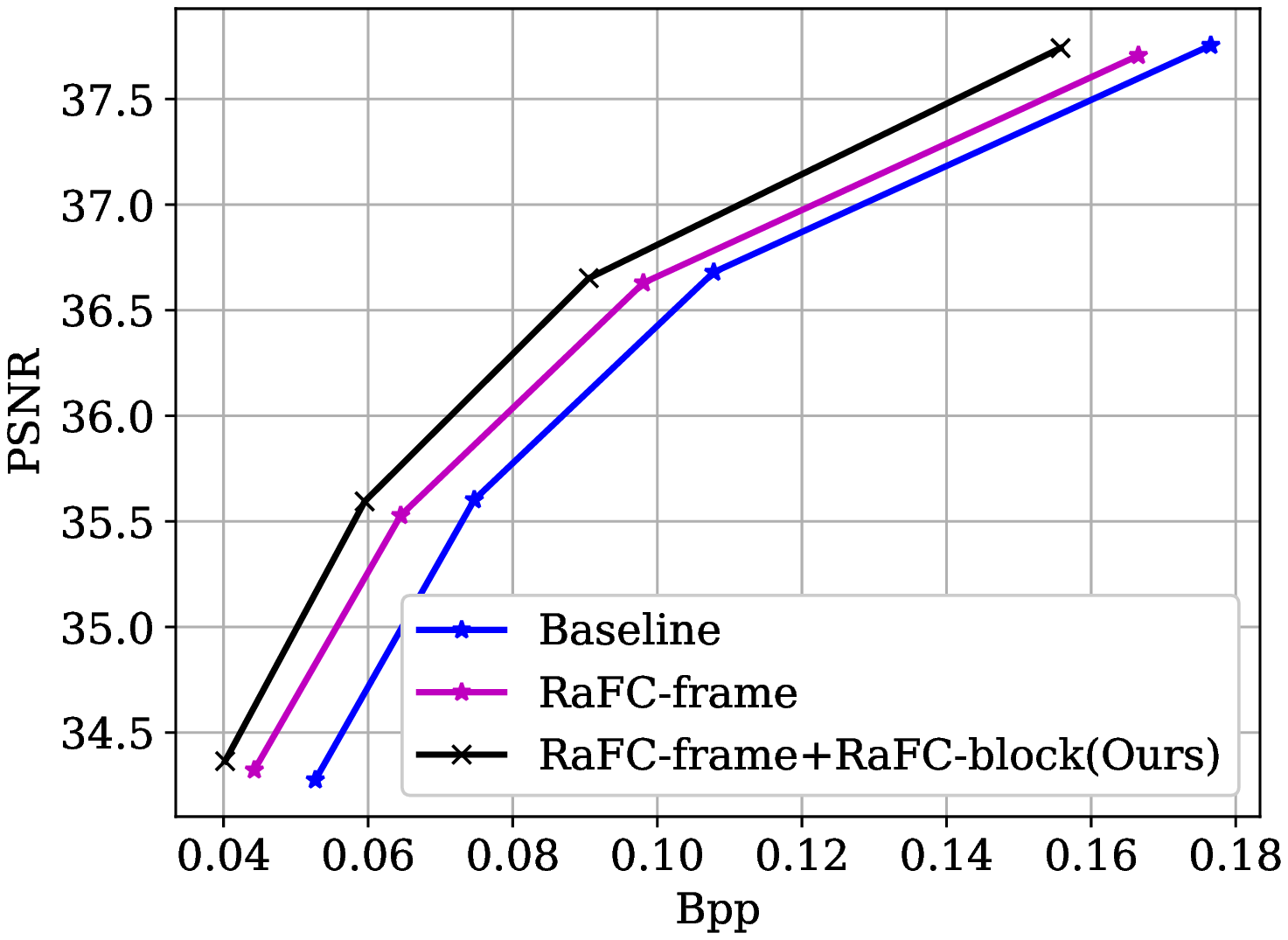}\label{fig:aba}}
\subfigure[Average PSNR(dB) over all predicted frames (i.e., $\bar{X}_t$'s) and the percentage of bits used to encode motion information over the total number of bits at different Bpps on the HEVC Class E dataset.
]{\includegraphics[height=1.40in]{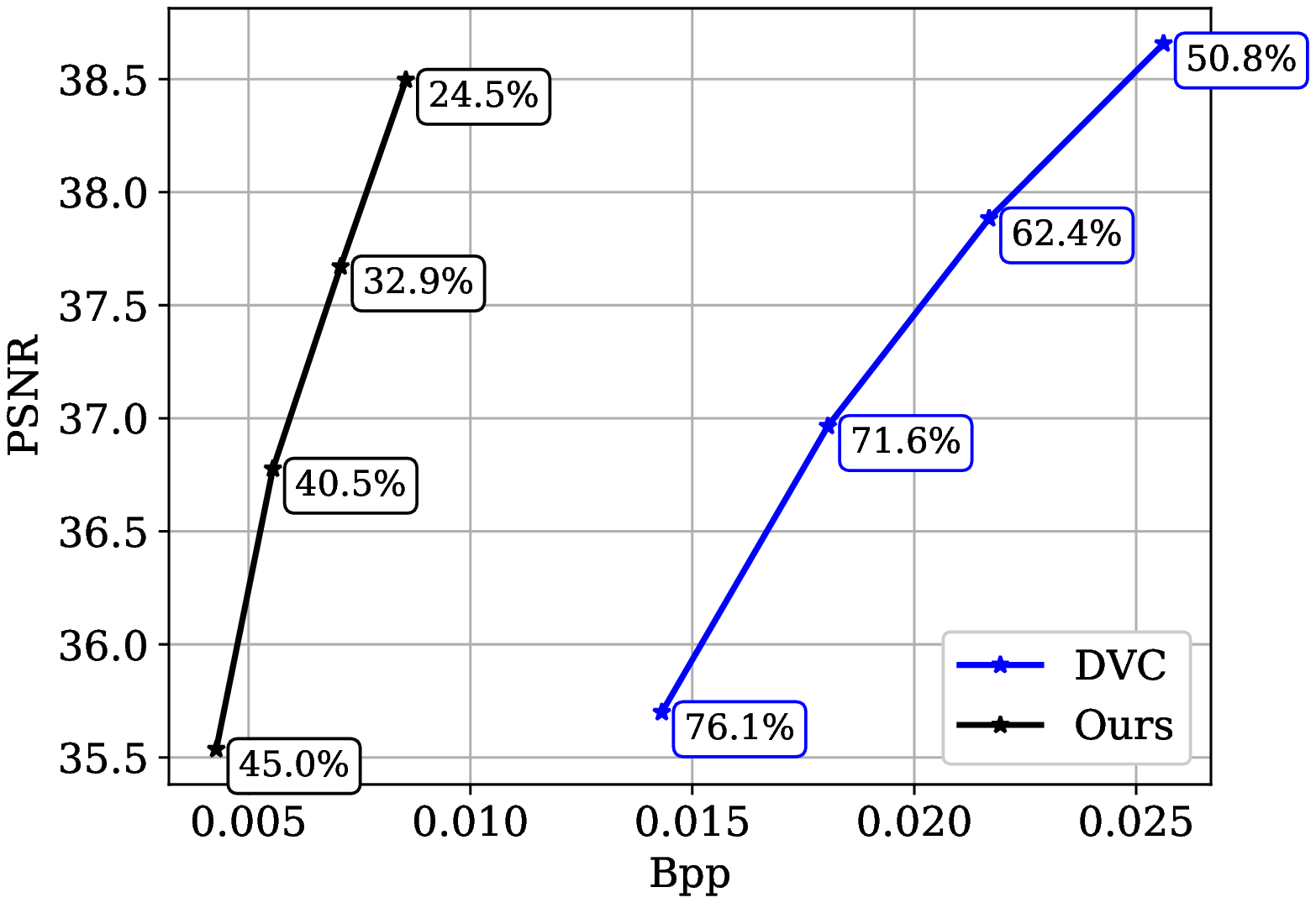}\label{fig:motionana}}
\caption{Ablation study and model analysis.}
\end{figure}

\subsection{Ablation Study and Model Analysis}

\textbf{Effectiveness of Different Components.} In order to verify the effectiveness of different components in our proposed method, we take the UVG dataset as an example to perform ablation study.
In this section, the hyperprior entropy model~\cite{balle2018variational} is used in all methods for fair comparison.
As shown in Figure~\ref{fig:aba}, our method RaFC-frame outperforms the baseline DVC algorithm and has achieved 0.5dB improvement when compared with DVC at 0.055bpp.
We also observe that our overall framework RaFC by using both RaFC-block scheme and RaFC-frame scheme achieves better result, which indicates that our overall framework combining RaFC-frame and RaFC-block can further improve the performance of RaFC-frame.
In other words, it is beneficial to choose the optimal resolution for both the optical flow maps and the corresponding motion representations.

\textbf{Model Analysis.} 
% Our proposed approach consists of multiple components including I-frame compression, motion compression and residual compression, while our method focuses on improving motion compression. 
In Figure~\ref{fig:motionana}, we take the HEVC Class E dataset as an example and show the average PSNR results over all predicted frames (i.e. $\bar{X}_t$'s) after motion compensation at different Bpps.
When compared with the flow coding method in DVC~\cite{lu2019dvc},
our overall RaFC framework can compress motion information in a much more effective way and save up to 70\% bits at the same PSNR when encoding motion information. 

Besides, we also report the percentage of bits used to encode motion information over the total number of bits for encoding both motion and residual information at different Bpps when using different $\lambda$ values.
And it is obvious that the percentage drops significantly when comparing our RaFC framework with the baseline DVC method, which indicates our RaFC framework uses less bits to encode flow information.

\begin{table}[!t]
\begin{minipage}[!t]{0.5\textwidth}
\begin{center}
    \setcaptionwidth{0.9\textwidth}
    \caption{Percentages of the selected optical flow map resolutions when using our RaFC-frame scheme at different $\lambda$ values.}
    \begin{tabular}{|l|c|c|}
    % \hline
    \hline
        & High resolution & Low resolution \\
        & (i.e., $\hat{V}_{t}^{1}$) & (i.e., $\hat{V}_{t}^{2}$) \\
    \hline
    \hline
        $\lambda = 256$ & 38.89\% & 61.11\%\\
        $\lambda = 512$ & 45.14\% & 54.86\%\\
        $\lambda = 1024$ & 57.64\% & 42.36\%\\
        $\lambda = 2048$ & 63.20\% & 36.80\%\\
    \hline
    % \hline
    \end{tabular}
    % }
    \label{tab:rafcfp}
\end{center}
\end{minipage}%
\begin{minipage}[!t]{0.5\textwidth}
\begin{center}
    \setcaptionwidth{0.9\textwidth}
    \caption{Percentages of the selected block resolutions when using our RaFC-block scheme at different $\lambda$ values.}
    \begin{tabular}{|l|c|c|c|}
    % \hline
    \hline
    Block    & $16\times16$ & $32\times32$ & $64\times64$ \\
    resolutions & (i.e., $\hat{m}_{t}^{1}$) & (i.e., $\hat{m}_{t}^{2}$) & (i.e., $\hat{m}_{t}^{3}$)\\
    \hline
    \hline
        $\lambda = 256$ & 0.98\% & 40.55\% & 58.46\%\\
        $\lambda = 512$ & 27.18\% & 36.69\% & 36.11\%\\
        $\lambda = 1024$ & 36.44\% & 32.27\% & 31.28\%\\
        $\lambda = 2048$ & 41.91\% & 31.02\% & 27.06\%\\
    \hline
    % \hline
    \end{tabular}
    % }
    \label{tab:rafcfbp}
\end{center}
\end{minipage}
    % \caption{Bits cost with different $\lambda$ on the HEVC dataset. ``Motion" means average bit per pixel (bpp) for encoding optical flow information while ``total" means average bpp for entire video coding. And the last column shows the proportion of bits for encoding motion information.}
    % \label{tab:motionbits}
\end{table}

\textbf{Resolutions Selection at Various Bit Rates.}
In our approach, we select the optimal resolution for the optical flow map in RaFC-frame or motion features in RaFC-block.
% flow from different resolutions (RaFC-Frame) or different motion features (RaFC-Block).
To investigate the effectiveness of our method, we provide the percentage of each selected resolution over the total number resolutions at various bit rates. From Table~\ref{tab:rafcfp} and Table~\ref{tab:rafcfbp}, we observe that low-resolution flow maps and large size blocks take a large portion at lower bit rates (\textit{i.e.,} when $\lambda$ is small). At higher bit rates (i.e., when $\lambda$ is large), it is more likely that our methods RaFC-frame and RaFC-block select high resolution flow maps and small block sizes, respectively. 
This observation is consist with the traditional video compression methods, where large size blocks are often preferred for motion estimation at low bit rates in order to save bits for motion coding.
% The reason is that the coding unit with large blocks can save more bitrates, which is important at low bitrate setting.

\begin{figure}[!t]
\centering
\subfigure[The 6th frame from the HEVC Class E dataset.]{\includegraphics[width=0.43\textwidth]{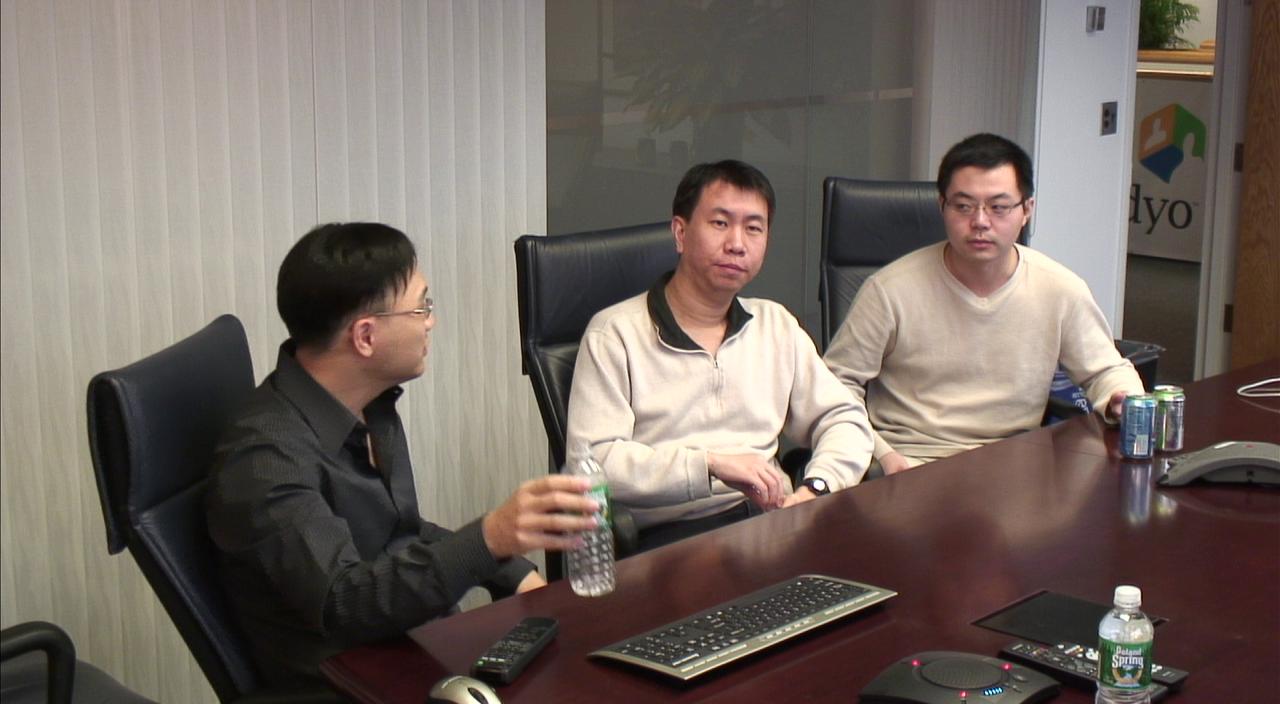}\label{fig:visinput}}%
\hfill
\subfigure[The reconstructed optical flow map and the corresponding block selection result by using our method RaFC-block.]{\includegraphics[width=0.43\textwidth]{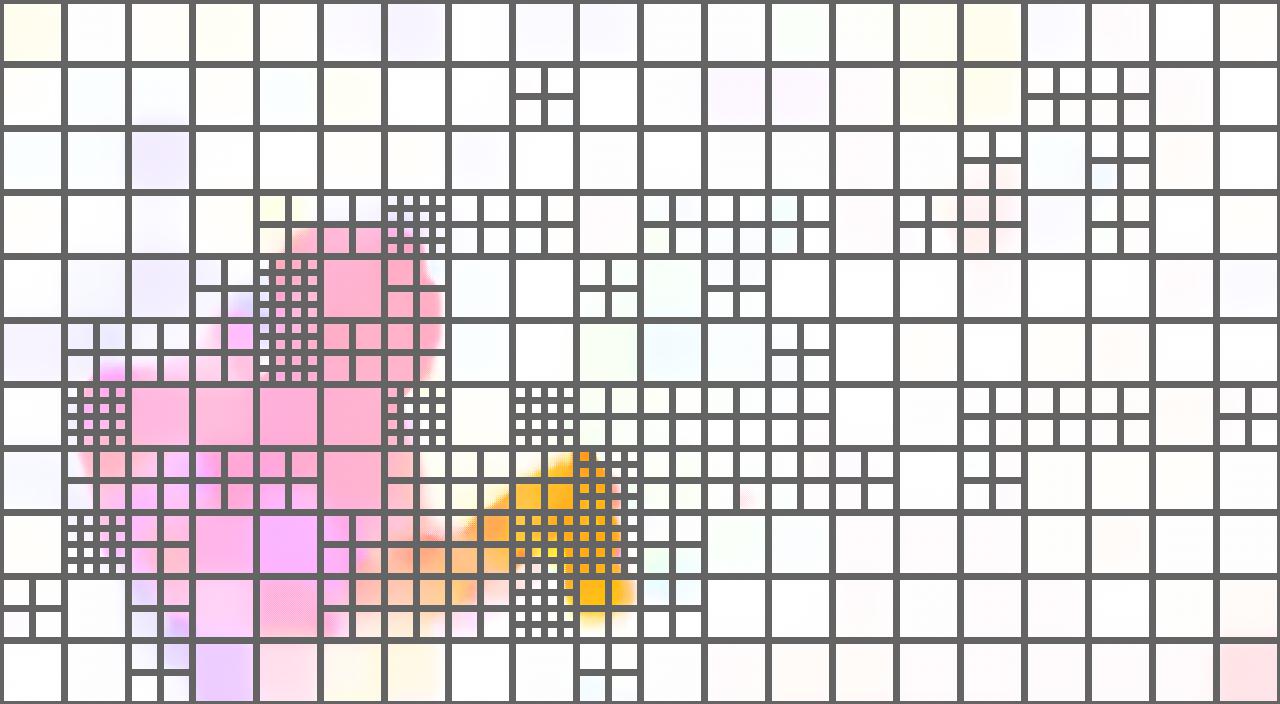}\label{fig:visflow}}
\caption{Visualization of the selected block resolutions by using our method RaFC-block.}
\label{fig:visblock}
\end{figure}

\textbf{Visualization of Selected Blocks.} In Figure~\ref{fig:visblock}, we visualize the selected blocks with different resolutions by using our method RaFC-block. Figure~\ref{fig:visinput} shows the 6th frame of the 1st video from the HEVC Class E dataset and Figure~\ref{fig:visflow} represents the reconstructed optical flow map of this frame and the corresponding block selection result by using our method RaFC-block. It can be observed that the small size blocks are often preferred from areas around the moving object boundaries and large size blocks are always preferred in the smooth areas.

\section{Conclusion}
In this work, we have proposed a Resolution-adaptive Flow Coding (RaFC) method to efficiently compress the motion information for video compression, which consists of two new schemes RaFC-frame at the frame-level and RaFC-block at the block-level.
Our method RaFC-frame can handle complex or simple motion patterns globally by automatically selecting the optimal resolutions from multi-scale flow maps, while our method RaFC-block can cope with different types of motion patterns locally by selecting the optimal resolutions of multi-scale motion features at each block. By performing comprehensive experiments on four benchmark datasets, we show that our RaFC framework outperforms the recent state-of-the-art deep learning based video compression methods. 
In our future work, we will use the proposed framework for encoding residual information and study more efficient block partitioning strategy.

\section*{Acknowledgement}
This work was supported by the National Key Research and Development Project of China (No. 2018AAA0101900). 
The work of Wanli Ouyang was supported by the Australian Medical Research Future Fund MRFAI000085.

\bibliographystyle{splncs04}
\bibliography{egbib}

\begin{thebibliography}{10}
\providecommand{\url}[1]{\texttt{#1}}
\providecommand{\urlprefix}{URL }
\providecommand{\doi}[1]{https://doi.org/#1}

\bibitem{UVGdataset}
Ultra video group test sequences. \url{http://ultravideo.cs.tut.fi}, accessed:
  2019-11-06

\bibitem{VTLdataset}
Video trace library. \url{http://trace.kom.aau.dk/yuv/index.html}, accessed:
  2019-11-06

\bibitem{agustsson2017soft}
Agustsson, E., Mentzer, F., Tschannen, M., Cavigelli, L., Timofte, R., Benini,
  L., Gool, L.V.: Soft-to-hard vector quantization for end-to-end learning
  compressible representations. In: Advances in Neural Information Processing
  Systems. pp. 1141--1151 (2017)

\bibitem{agustsson2019generative}
Agustsson, E., Tschannen, M., Mentzer, F., Timofte, R., Gool, L.V.: Generative
  adversarial networks for extreme learned image compression. In: Proceedings
  of the IEEE International Conference on Computer Vision. pp. 221--231 (2019)

\bibitem{balle2016end}
Ball{\'e}, J., Laparra, V., Simoncelli, E.P.: End-to-end optimized image
  compression. International Conference on Learning Representations (ICLR)
  (2017)

\bibitem{balle2018variational}
Ball{\'e}, J., Minnen, D., Singh, S., Hwang, S.J., Johnston, N.: Variational
  image compression with a scale hyperprior. International Conference on
  Learning Representations (ICLR)  (2018)

\bibitem{bellard2015bpg}
Bellard, F.: Bpg image format. URL https://bellard. org/bpg  (2015)

\bibitem{chen2019learning}
Chen, Z., He, T., Jin, X., Wu, F.: Learning for video compression. IEEE
  Transactions on Circuits and Systems for Video Technology  (2019)

\bibitem{abdelaziz2019neural}
Djelouah, A., Campos, J., Schaub-Meyer, S., Schroers, C.: Neural inter-frame
  compression for video coding. In: Proceedings of the IEEE International
  Conference on Computer Vision. pp. 6421--6429 (2019)

\bibitem{dosovitskiy2015flownet}
Dosovitskiy, A., Fischer, P., Ilg, E., Hausser, P., Hazirbas, C., Golkov, V.,
  Van Der~Smagt, P., Cremers, D., Brox, T.: Flownet: Learning optical flow with
  convolutional networks. In: Proceedings of the IEEE international conference
  on computer vision. pp. 2758--2766 (2015)

\bibitem{guo2020content}
Guo, L., Chunlei, C., Xiaoyun, Z., Li, C., Wanli, O., Dong, X., Zhiyong, G.:
  Content adaptive and error propagation aware deep video compression.
  Proceedings of the European Conference on Computer Vision (ECCV)  (2020)

\bibitem{habibian2019video}
Habibian, A., Rozendaal, T.v., Tomczak, J.M., Cohen, T.S.: Video compression
  with rate-distortion autoencoders. In: Proceedings of the IEEE International
  Conference on Computer Vision. pp. 7033--7042 (2019)

\bibitem{hui2018liteflownet}
Hui, T.W., Tang, X., Change~Loy, C.: Liteflownet: A lightweight convolutional
  neural network for optical flow estimation. In: Proceedings of the IEEE
  Conference on Computer Vision and Pattern Recognition. pp. 8981--8989 (2018)

\bibitem{johnston2018improved}
Johnston, N., Vincent, D., Minnen, D., Covell, M., Singh, S., Chinen, T.,
  Jin~Hwang, S., Shor, J., Toderici, G.: Improved lossy image compression with
  priming and spatially adaptive bit rates for recurrent networks. In:
  Proceedings of the IEEE Conference on Computer Vision and Pattern
  Recognition. pp. 4385--4393 (2018)

\bibitem{kingma2014adam}
Kingma, D.P., Ba, J.: Adam: A method for stochastic optimization. International
  Conference for Learning Representations  (2015)

\bibitem{li2018learning}
Li, M., Zuo, W., Gu, S., Zhao, D., Zhang, D.: Learning convolutional networks
  for content-weighted image compression. In: Proceedings of the IEEE
  Conference on Computer Vision and Pattern Recognition. pp. 3214--3223 (2018)

\bibitem{lu2019dvc}
Lu, G., Ouyang, W., Xu, D., Zhang, X., Cai, C., Gao, Z.: Dvc: An end-to-end
  deep video compression framework. In: Proceedings of the IEEE Conference on
  Computer Vision and Pattern Recognition. pp. 11006--11015 (2019)

\bibitem{lu2020anendtoend}
Lu, G., Zhang, X., Ouyang, W., Chen, L., Gao, Z., Xu, D.: An end-to-end
  learning framework for video compression. IEEE Transactions on Pattern
  Analysis and Machine Intelligence  \textbf{in Press}, ~1--1.
  \doi{10.1109/TPAMI.2020.2988453}

\bibitem{minnen2018joint}
Minnen, D., Ball{\'e}, J., Toderici, G.D.: Joint autoregressive and
  hierarchical priors for learned image compression. In: Advances in Neural
  Information Processing Systems. pp. 10771--10780 (2018)

\bibitem{ranjan2017optical}
Ranjan, A., Black, M.J.: Optical flow estimation using a spatial pyramid
  network. In: Proceedings of the IEEE Conference on Computer Vision and
  Pattern Recognition. pp. 4161--4170 (2017)

\bibitem{rippel2017real}
Rippel, O., Bourdev, L.: Real-time adaptive image compression. In: Proceedings
  of the 34th International Conference on Machine Learning-Volume 70. pp.
  2922--2930. JMLR. org (2017)

\bibitem{rippel2019learned}
Rippel, O., Nair, S., Lew, C., Branson, S., Anderson, A.G., Bourdev, L.:
  Learned video compression. In: Proceedings of the IEEE International
  Conference on Computer Vision. pp. 3454--3463 (2019)

\bibitem{sullivan2012overview}
Sullivan, G.J., Ohm, J.R., Han, W.J., Wiegand, T.: Overview of the high
  efficiency video coding (hevc) standard. IEEE Transactions on circuits and
  systems for video technology  \textbf{22}(12),  1649--1668 (2012)

\bibitem{taubman2002jpeg2000}
Taubman, D.S., Marcellin, M.W.: Jpeg2000: Standard for interactive imaging.
  Proceedings of the IEEE  \textbf{90}(8),  1336--1357 (2002)

\bibitem{theis2017lossy}
Theis, L., Shi, W., Cunningham, A., Husz{\'a}r, F.: Lossy image compression
  with compressive autoencoders. International Conference for Learning
  Representations  (2017)

\bibitem{toderici2015variable}
Toderici, G., O'Malley, S.M., Hwang, S.J., Vincent, D., Minnen, D., Baluja, S.,
  Covell, M., Sukthankar, R.: Variable rate image compression with recurrent
  neural networks. International Conference for Learning Representations
  (2017)

\bibitem{toderici2017full}
Toderici, G., Vincent, D., Johnston, N., Jin~Hwang, S., Minnen, D., Shor, J.,
  Covell, M.: Full resolution image compression with recurrent neural networks.
  In: Proceedings of the IEEE Conference on Computer Vision and Pattern
  Recognition. pp. 5306--5314 (2017)

\bibitem{tsai2018learning}
Tsai, Y.H., Liu, M.Y., Sun, D., Yang, M.H., Kautz, J.: Learning binary residual
  representations for domain-specific video streaming. In: Thirty-Second AAAI
  Conference on Artificial Intelligence (2018)

\bibitem{wallace1992jpeg}
Wallace, G.K.: The jpeg still picture compression standard. IEEE transactions
  on consumer electronics  \textbf{38}(1),  xviii--xxxiv (1992)

\bibitem{wang2016mcl}
Wang, H., Gan, W., Hu, S., Lin, J.Y., Jin, L., Song, L., Wang, P.,
  Katsavounidis, I., Aaron, A., Kuo, C.C.J.: Mcl-jcv: a jnd-based h. 264/avc
  video quality assessment dataset. In: 2016 IEEE International Conference on
  Image Processing (ICIP). pp. 1509--1513. IEEE (2016)

\bibitem{wang2003multiscale}
Wang, Z., Simoncelli, E.P., Bovik, A.C.: Multiscale structural similarity for
  image quality assessment. In: The Thrity-Seventh Asilomar Conference on
  Signals, Systems \& Computers, 2003. vol.~2, pp. 1398--1402. Ieee (2003)

\bibitem{wiegand2003overview}
Wiegand, T., Sullivan, G.J., Bjontegaard, G., Luthra, A.: Overview of the h.
  264/avc video coding standard. IEEE Transactions on circuits and systems for
  video technology  \textbf{13}(7),  560--576 (2003)

\bibitem{wu2018video}
Wu, C.Y., Singhal, N., Krahenbuhl, P.: Video compression through image
  interpolation. In: Proceedings of the European Conference on Computer Vision
  (ECCV). pp. 416--431 (2018)

\bibitem{xu2018reduction}
{Xu}, M., {Li}, T., {Wang}, Z., {Deng}, X., {Yang}, R., {Guan}, Z.: Reducing
  complexity of hevc: A deep learning approach. IEEE Transactions on Image
  Processing  \textbf{27}(10),  5044--5059 (2018)

\bibitem{xue2019video}
Xue, T., Chen, B., Wu, J., Wei, D., Freeman, W.T.: Video enhancement with
  task-oriented flow. International Journal of Computer Vision
  \textbf{127}(8),  1106--1125 (2019)

\end{thebibliography}

% \title{Improving Deep Video Compression by Resolution-adaptive Flow Coding (Supplementary Materials)} % Replace with your title

%   \titlerunning{Improving Deep Video Compression by Resolution-adaptive Flow Coding}
%   % If the paper title is too long for the running head, you can set
%   % an abbreviated paper title here
%   %
%   \author{}
%   %
%   \authorrunning{Zhihao Hu et al.}
%   % First names are abbreviated in the running head.
%   % If there are more than two authors, 'et al.' is used.
%   %
%   \institute{}
% \maketitle
\setcounter{section}{0}

\renewcommand\thesection{\Alph{section}}
\section{Results on the HEVC Class B, Class C and Class D datasets}
% \begin{figure*}[t]
% \begin{center}
% \includegraphics[width=7.0in]{figures/CDresult.pdf}
% \end{center}
%    \caption{Results of different methods on the HEVC Class C and Class D datasets.}
% \label{fig:result}
% \end{figure*}

\begin{figure}[t]
  \centering
  \begin{minipage}[c]{0.5\textwidth}
    \centering
    \includegraphics[height=1.7in]{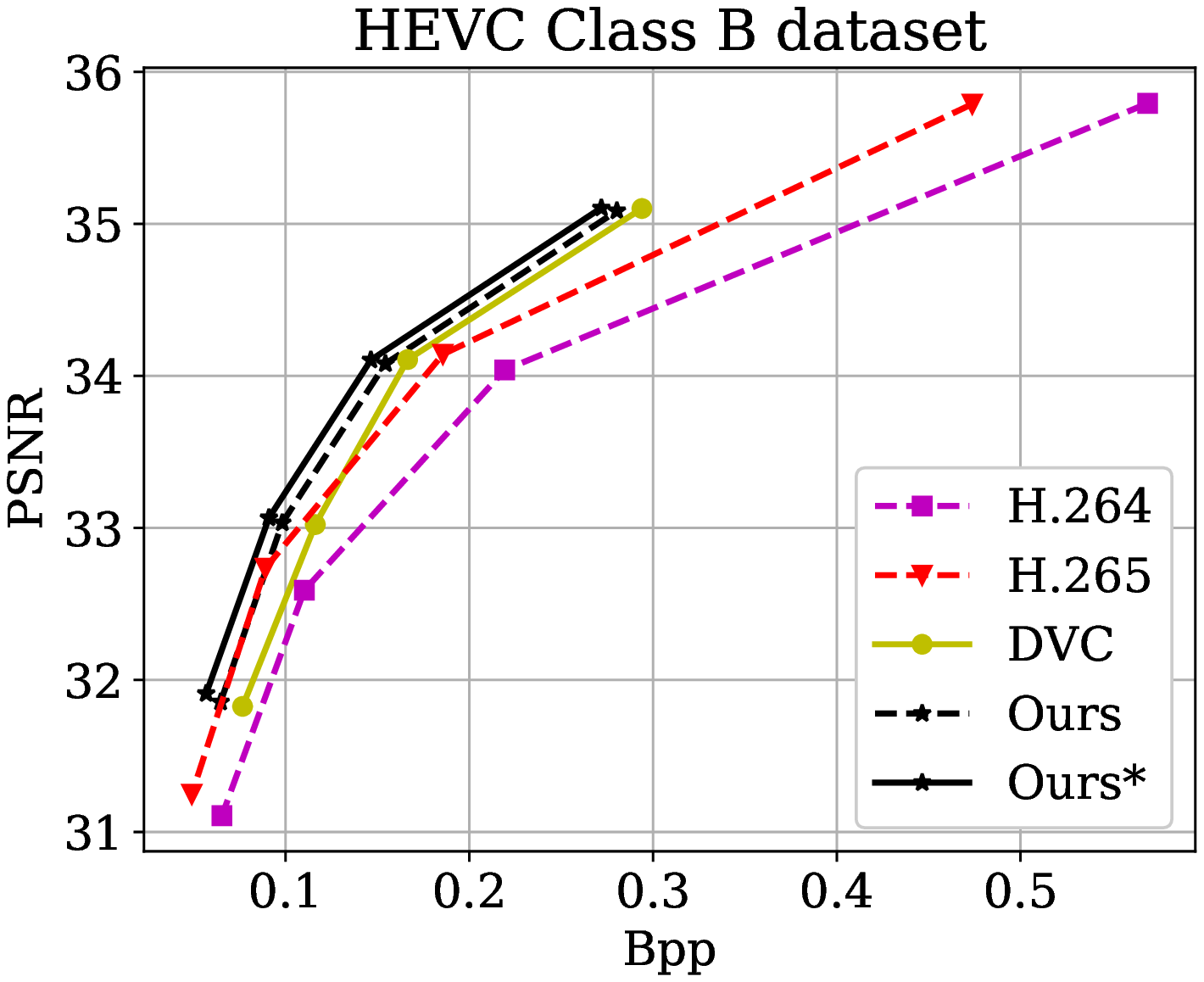}
  \end{minipage}%
  \begin{minipage}[c]{0.5\textwidth}
  \centering
    \includegraphics[height=1.7in]{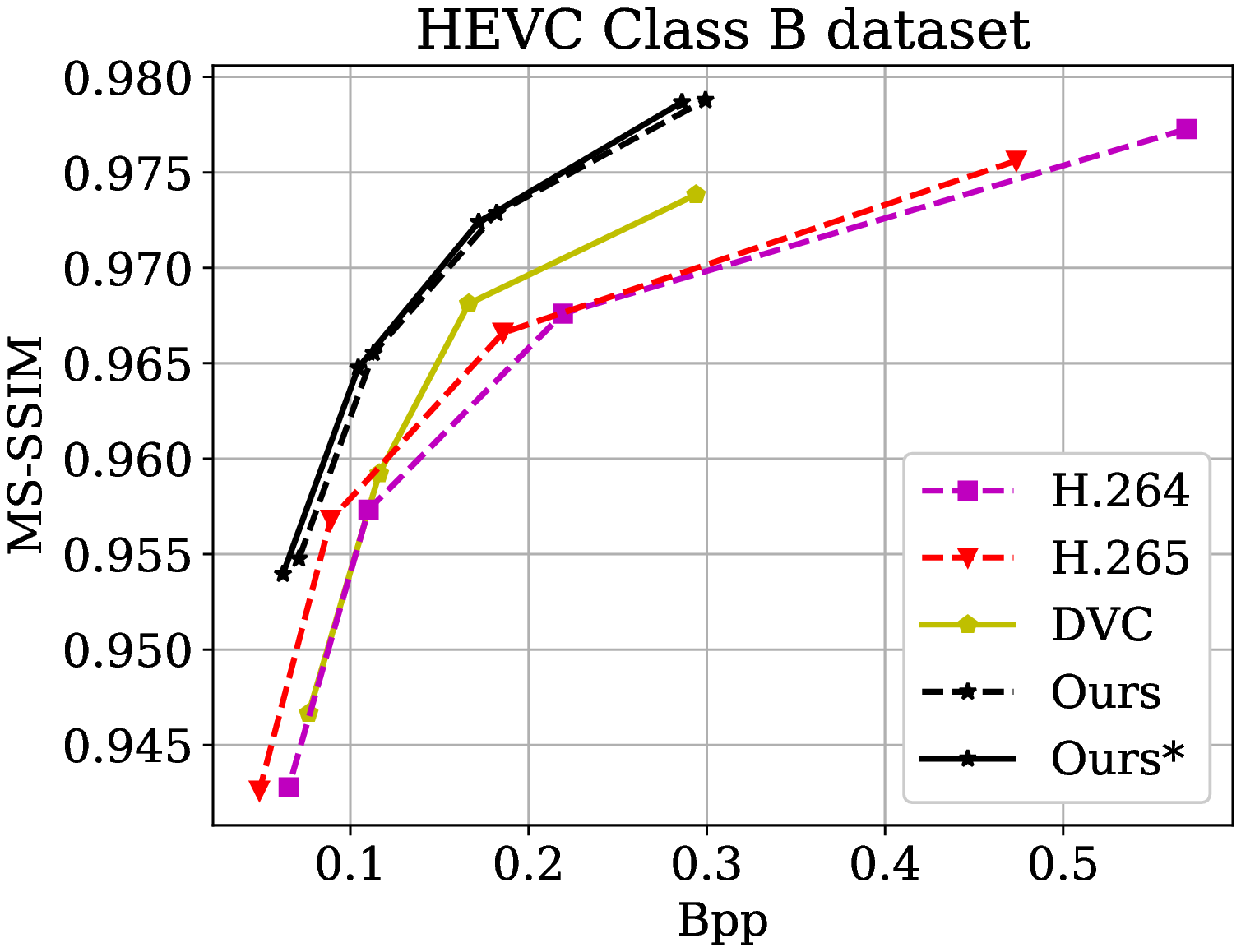}
  \end{minipage}
  \begin{minipage}[c]{0.5\textwidth}
    \centering
    \includegraphics[height=1.7in]{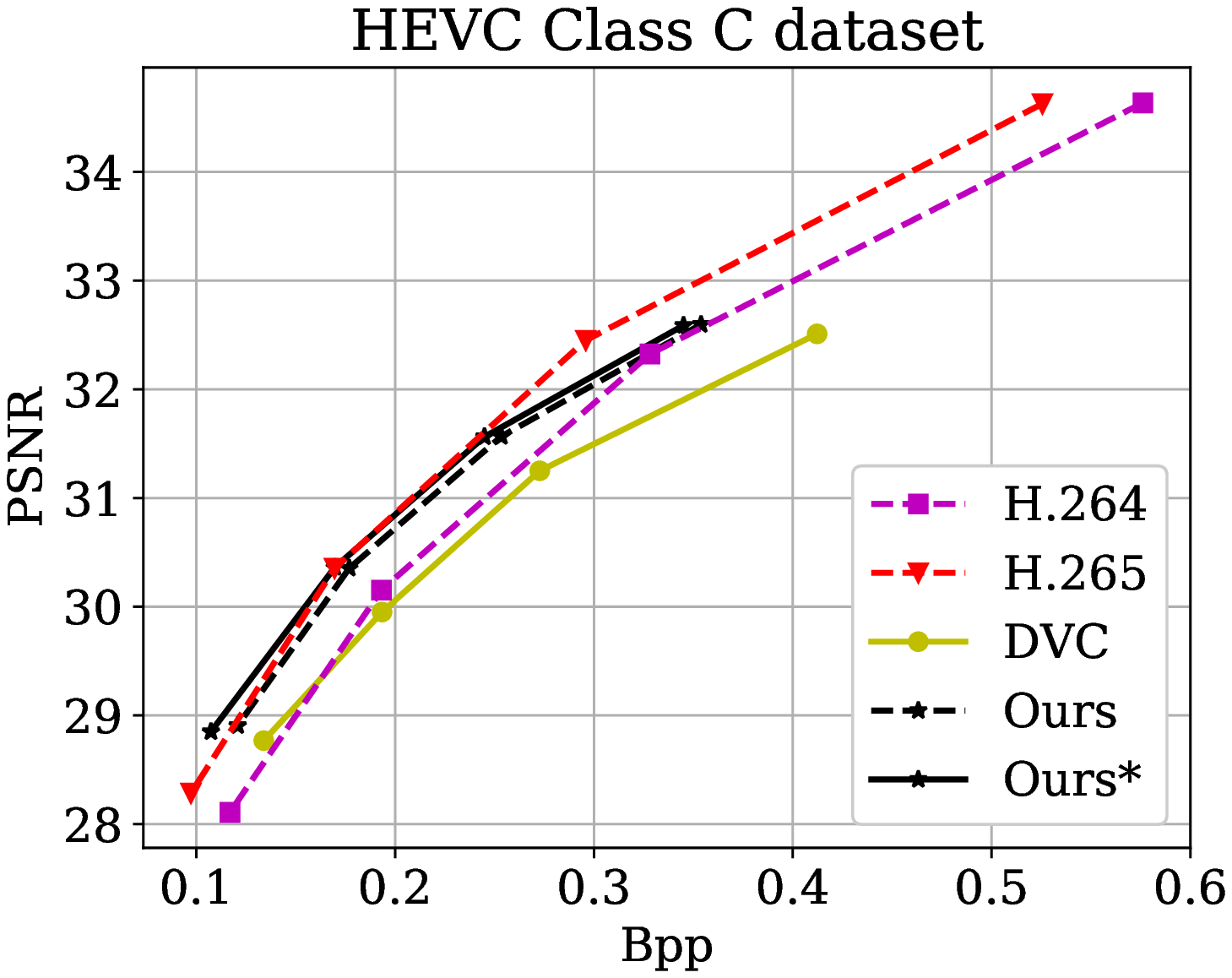}
  \end{minipage}%
  \begin{minipage}[c]{0.5\textwidth}
  \centering
    \includegraphics[height=1.7in]{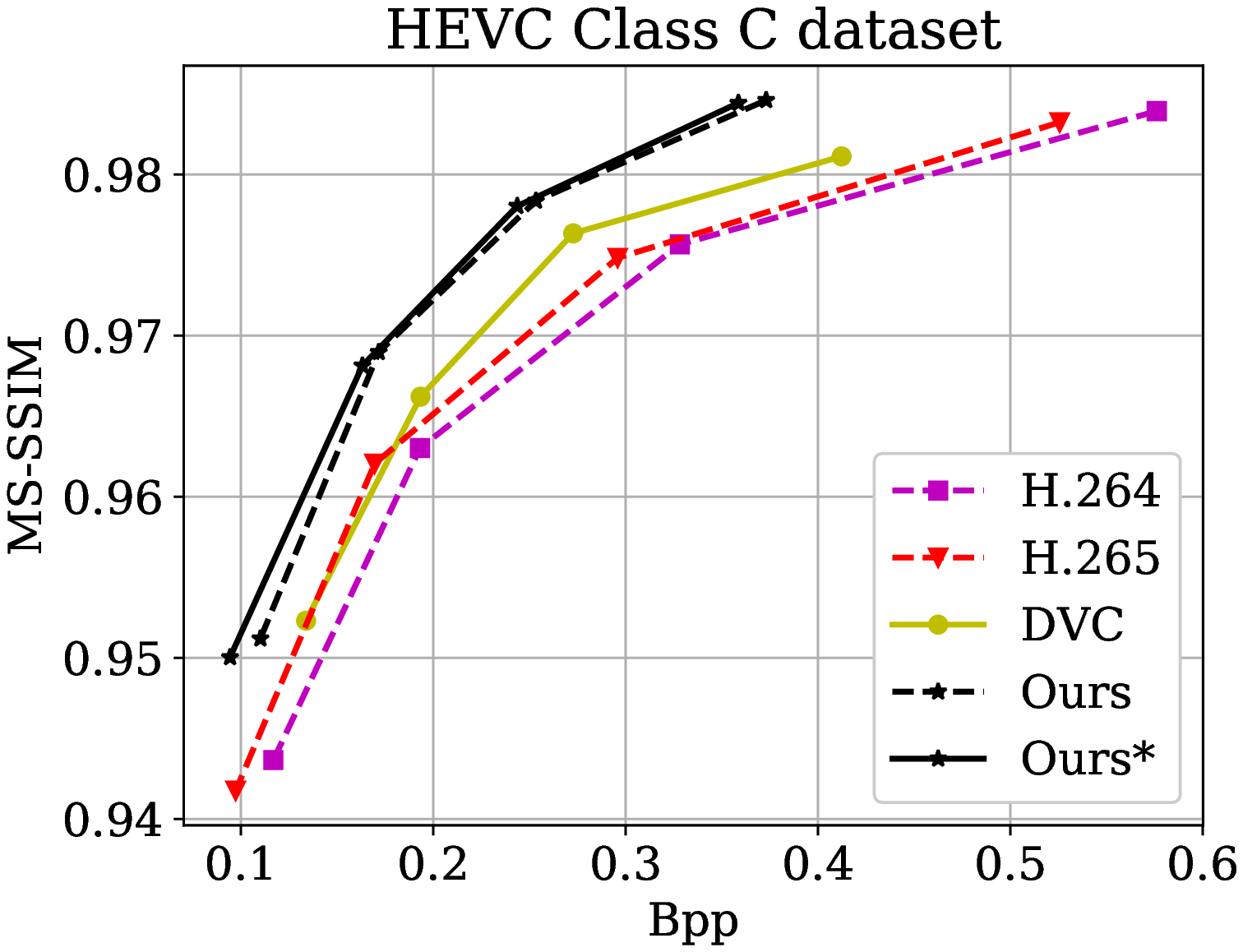}
  \end{minipage}
  \begin{minipage}[c]{0.5\textwidth}
    \centering
    \includegraphics[height=1.7in]{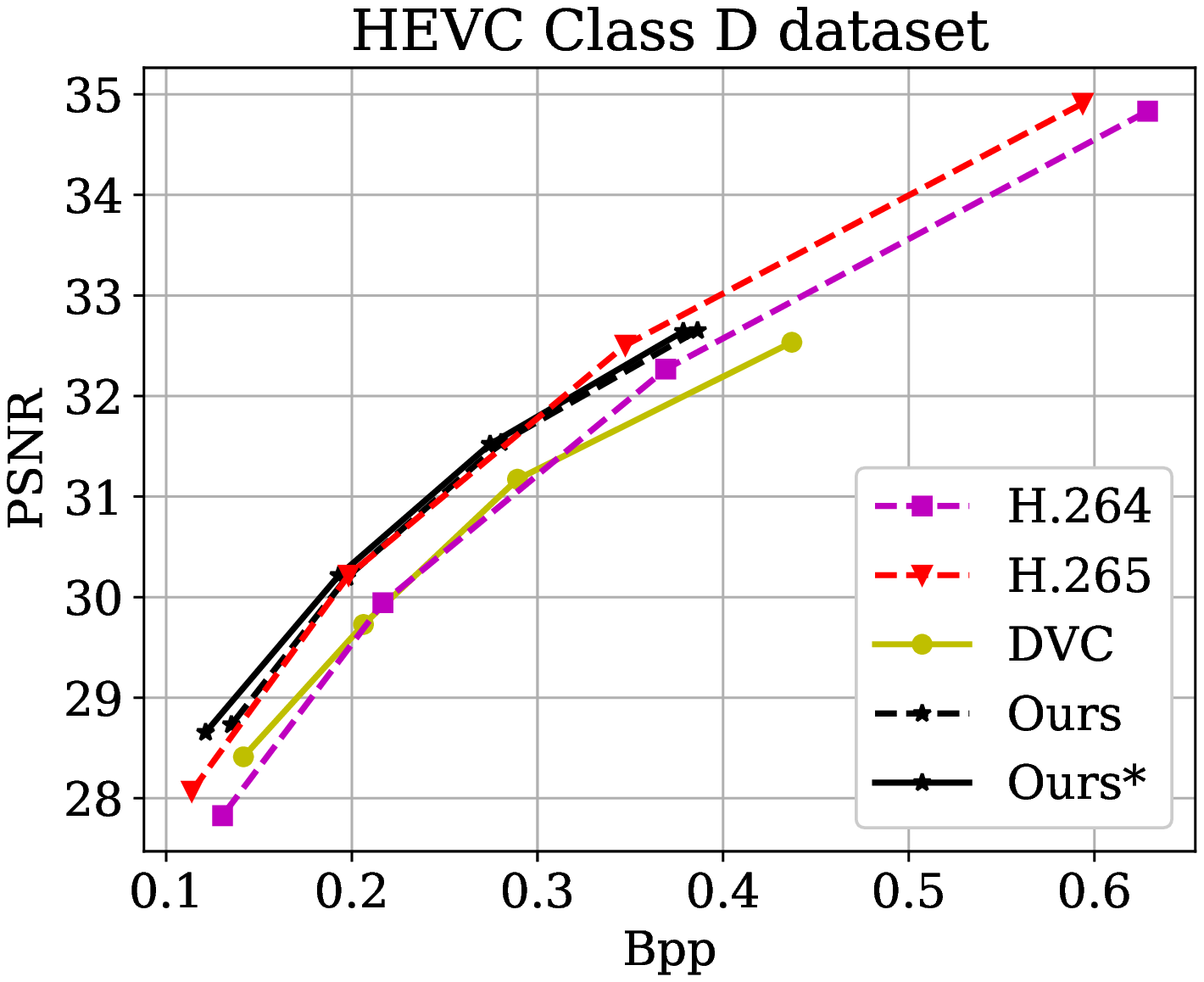}
  \end{minipage}%
  \begin{minipage}[c]{0.5\textwidth}
  \centering
    \includegraphics[height=1.7in]{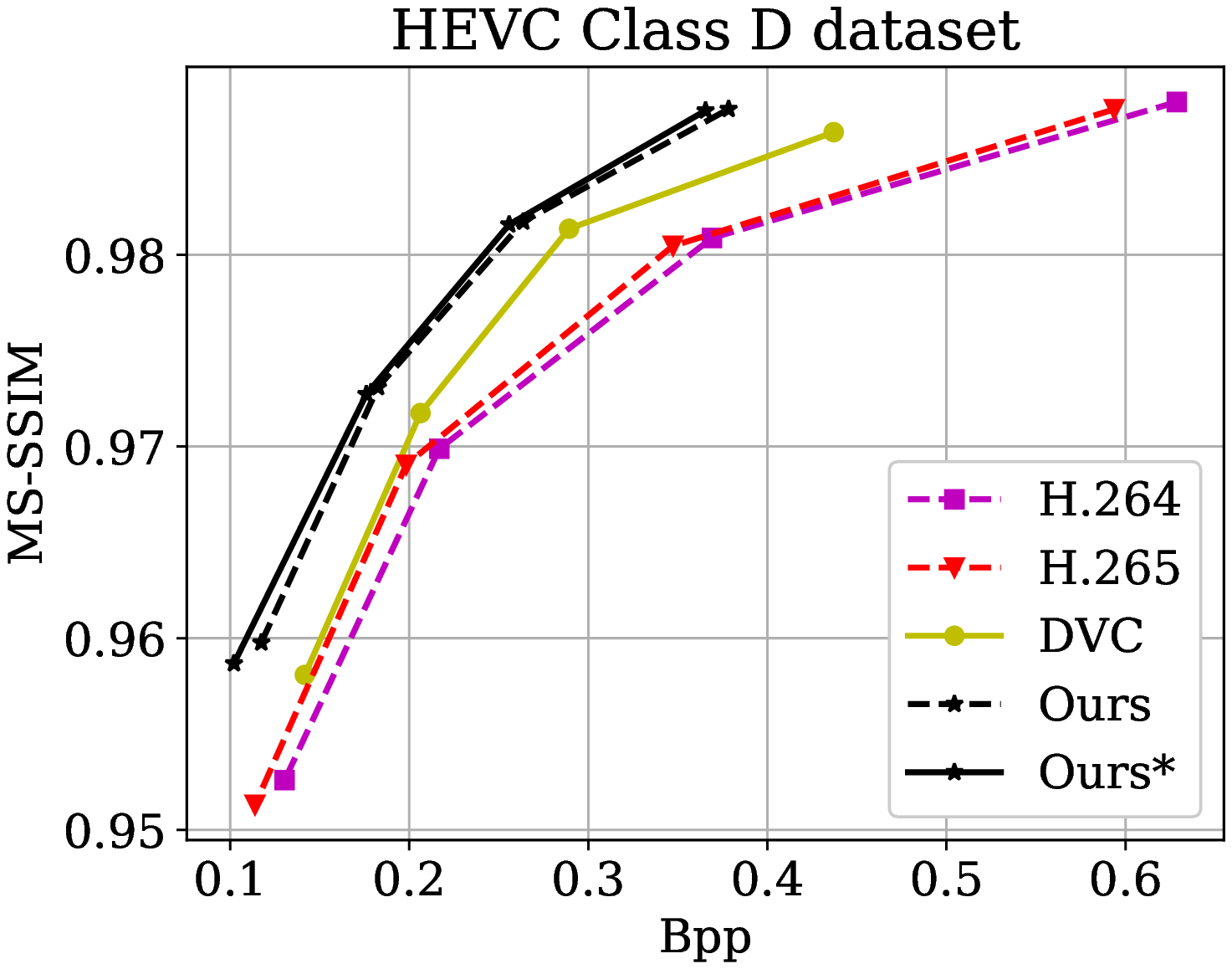}
  \end{minipage}
    \caption{The experimental results of different methods on the HEVC Class B, Class C and Class D datasets.}
  \label{fig:result}
\end{figure}

In the main paper, the results of different methods on four datasets HEVC Class E, VTL, UVG and MCL-JVC are provided and we have also mentioned that the results on the HEVC Class B, Class C and Class D datasets will be provided in the \textit{supplementary material}. Figure~\ref{fig:result} reports the results of different methods on the HEVC Class B, Class C and Class D datasets. When compared with H.265, we observe that our method still achieves competitive results at low bit rates in terms of PSNR. Additionally, our method outperforms all other baseline methods in terms of MS-SSIM.

% \section{Results on the MCL-JVC dataset}
% % \begin{figure*}[t]
% % \begin{center}
% % \includegraphics[width=7.0in]{figures/MCLresult.pdf}
% % \end{center}
% %    \caption{Results of different methods on the MCL-JVC dataset.}
% % \label{fig:mclresult}
% % \end{figure*}

% In Figure~\ref{fig:mclresult}, we compare our method with the baseline methods on the MCL\_JVC dataset (due to space limitation, the results cannot be provided in the main paper). Our method outperforms H.265 at almost all the bit rates in terms of both PSNR and MS-SSIM. When compared with AD\_ICCV19 \cite{abdelaziz2019neural}, our method also achieves about 1.0dB improvement at low bit rates and about 0.5dB improvement at high bit rates. In addition, our method outperforms other deep learning based baseline methods at all bit rates.

\section{The Command Line for H.264 and H.265}

We follow the setting in \cite{lu2019dvc} to use the FFmpeg to generate the compressed videos from H.264 and H.265 with the \textit{default} mode.
For the uncompressed video \textit{A.yuv} with the resolution of $W \times H$, the command line for generating compressed video \textit{output.mkv} by using the H.264  is provided as follows,

\textit{ffmpeg -pix\_fmt yuv420p -s WxH -r FR -i A.yuv -vframes N -c:v libx264 -tune zerolatency -crf Q -g GoP -sc\_threshold 0 output.mkv}

And the command line for H.265 is provided as follows,

\textit{ffmpeg -pix\_fmt yuv420p -s WxH -r FR -i A.yuv -vframes N -c:v libx265 -tune zerolatency -x265-params ``crf=Q:keyint=GoP:verbose=1" output.mkv}

where \textit{FR, N, Q, GoP} represent the frame rate, the number of frames to be encoded, the quality and the GoP size. \textit{Q} is set as 19, 23, 27, 31. \textit{GoP} is set as 10 for the HEVC dataset and 12 for other datasets.

\section{Results of Our Method and H.265 Using Variable GoP Sizes}

\begin{figure}[h]
  \centering
  \begin{minipage}[c]{0.5\textwidth}
    \centering
    \includegraphics[height=1.75in]{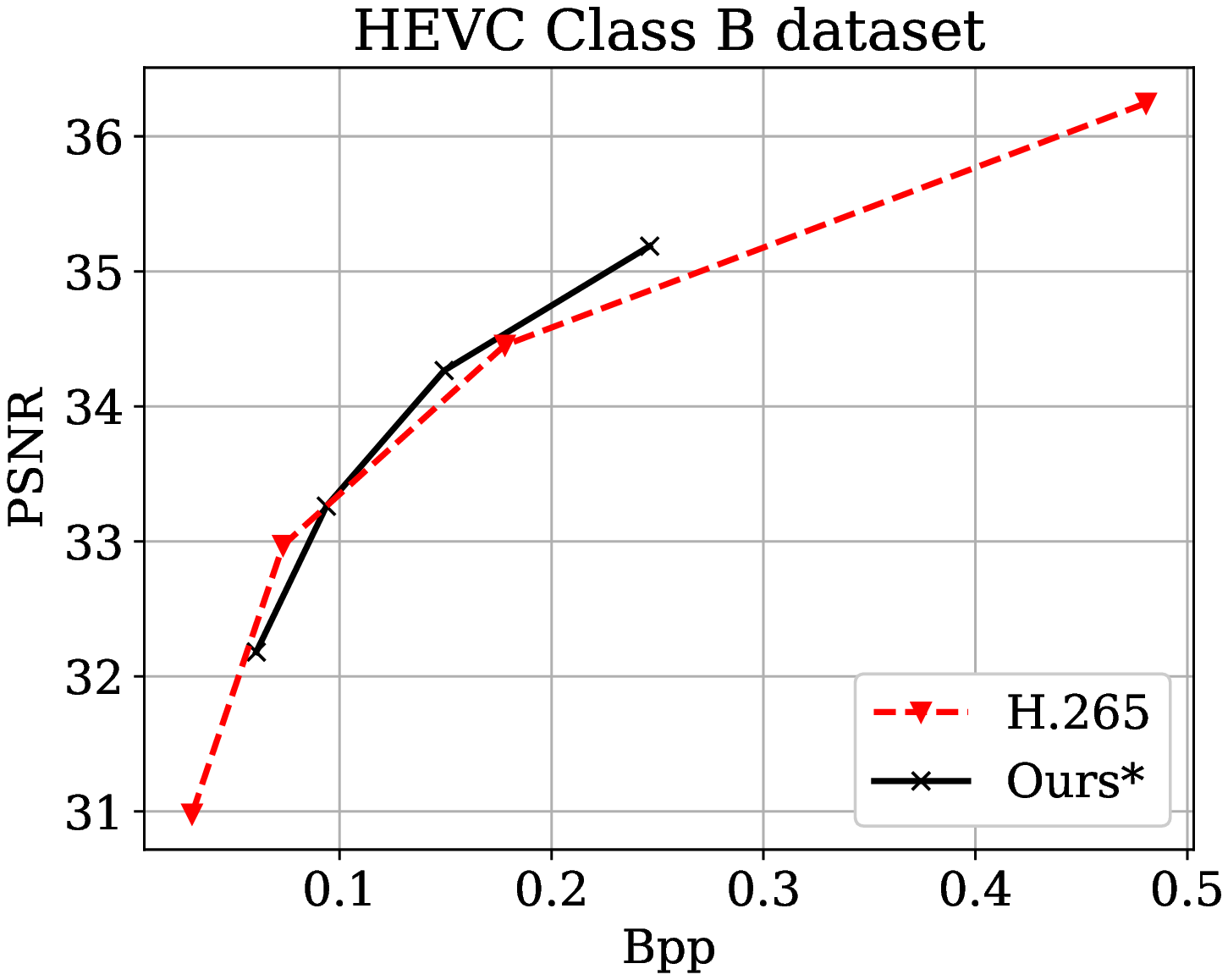}
  \end{minipage}%
  \begin{minipage}[c]{0.5\textwidth}
    \centering
    \includegraphics[height=1.75in]{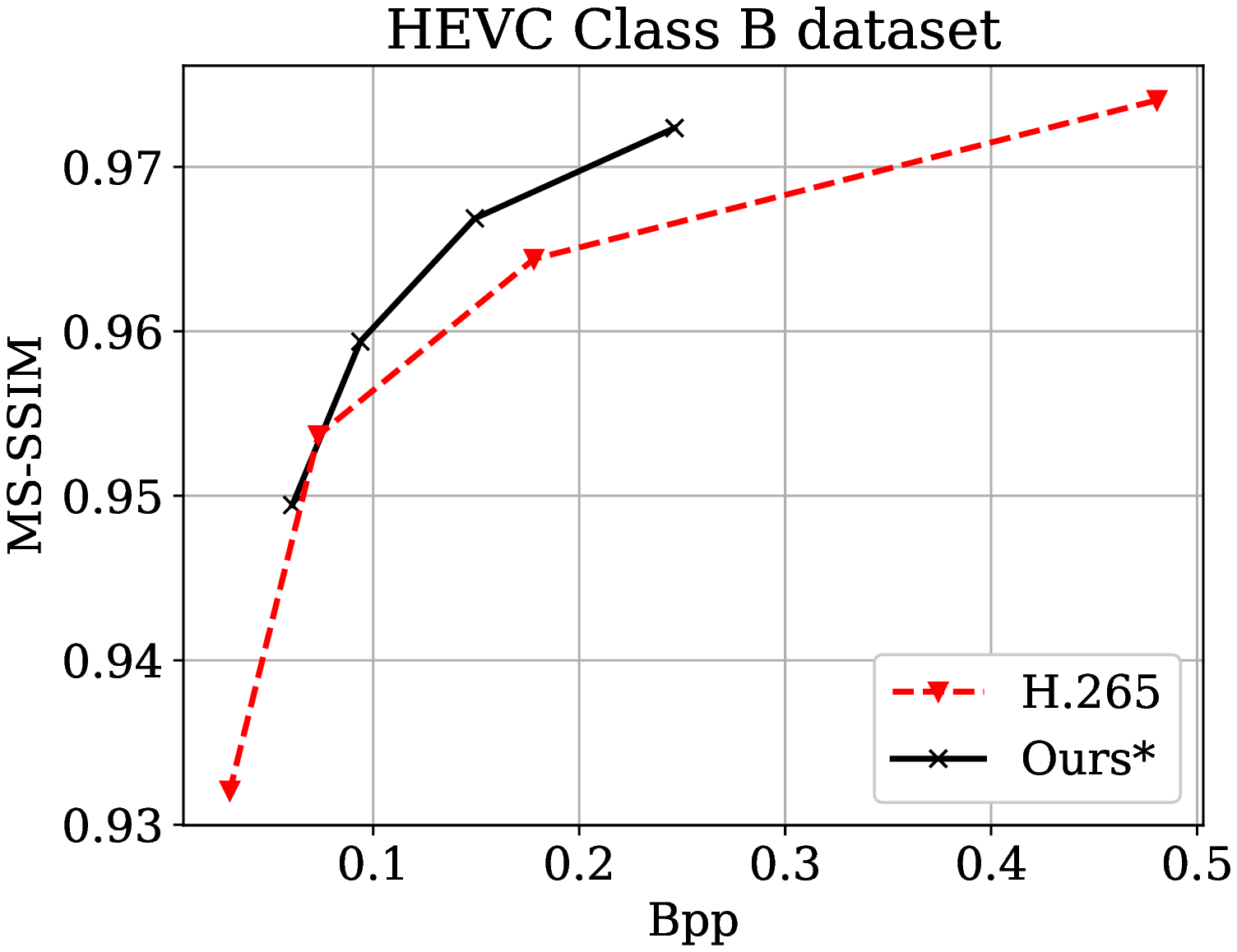}
  \end{minipage}
  \begin{minipage}[c]{0.5\textwidth}
  \centering
    \includegraphics[height=1.75in]{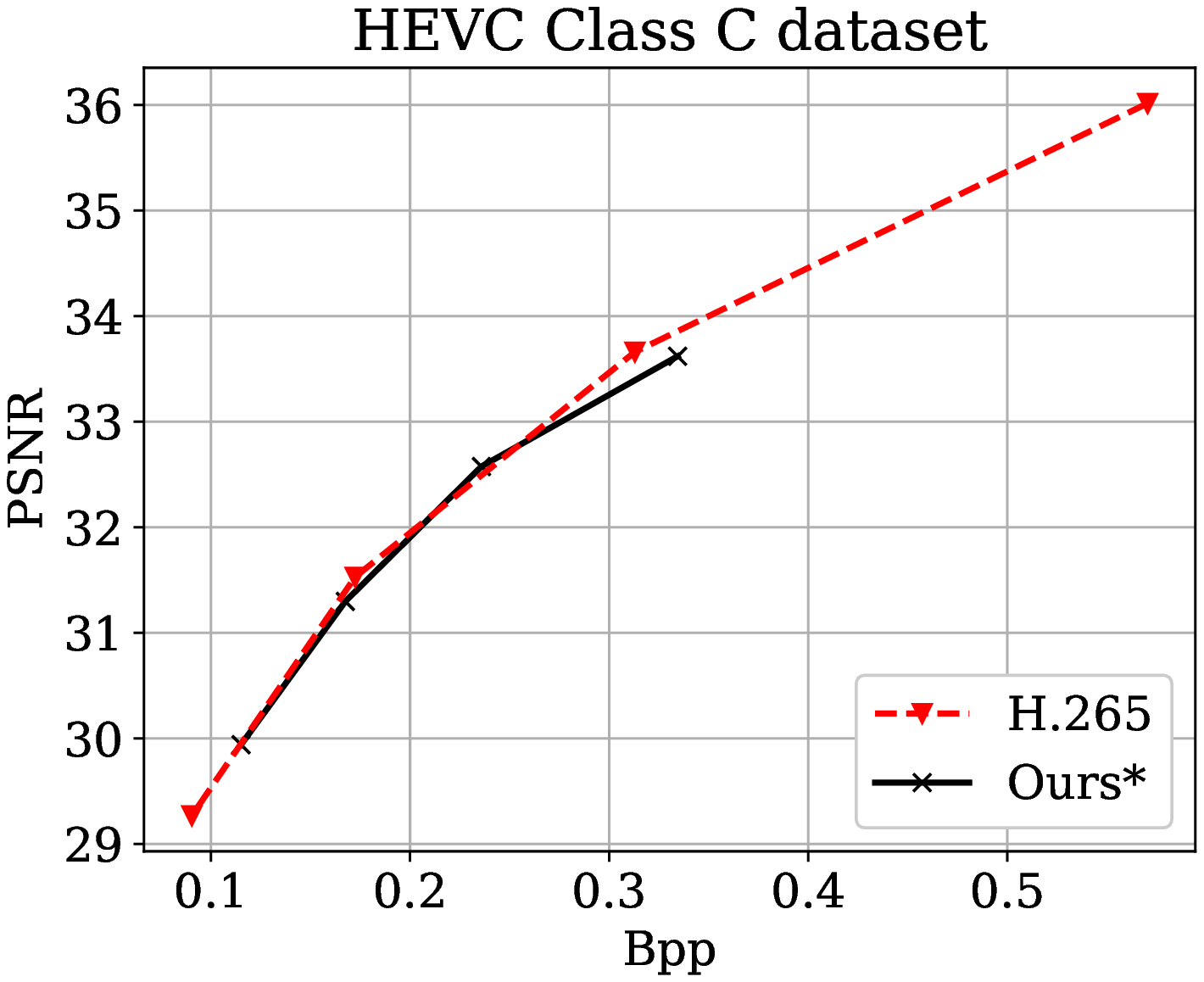}
  \end{minipage}%
  \begin{minipage}[c]{0.5\textwidth}
  \centering
    \includegraphics[height=1.75in]{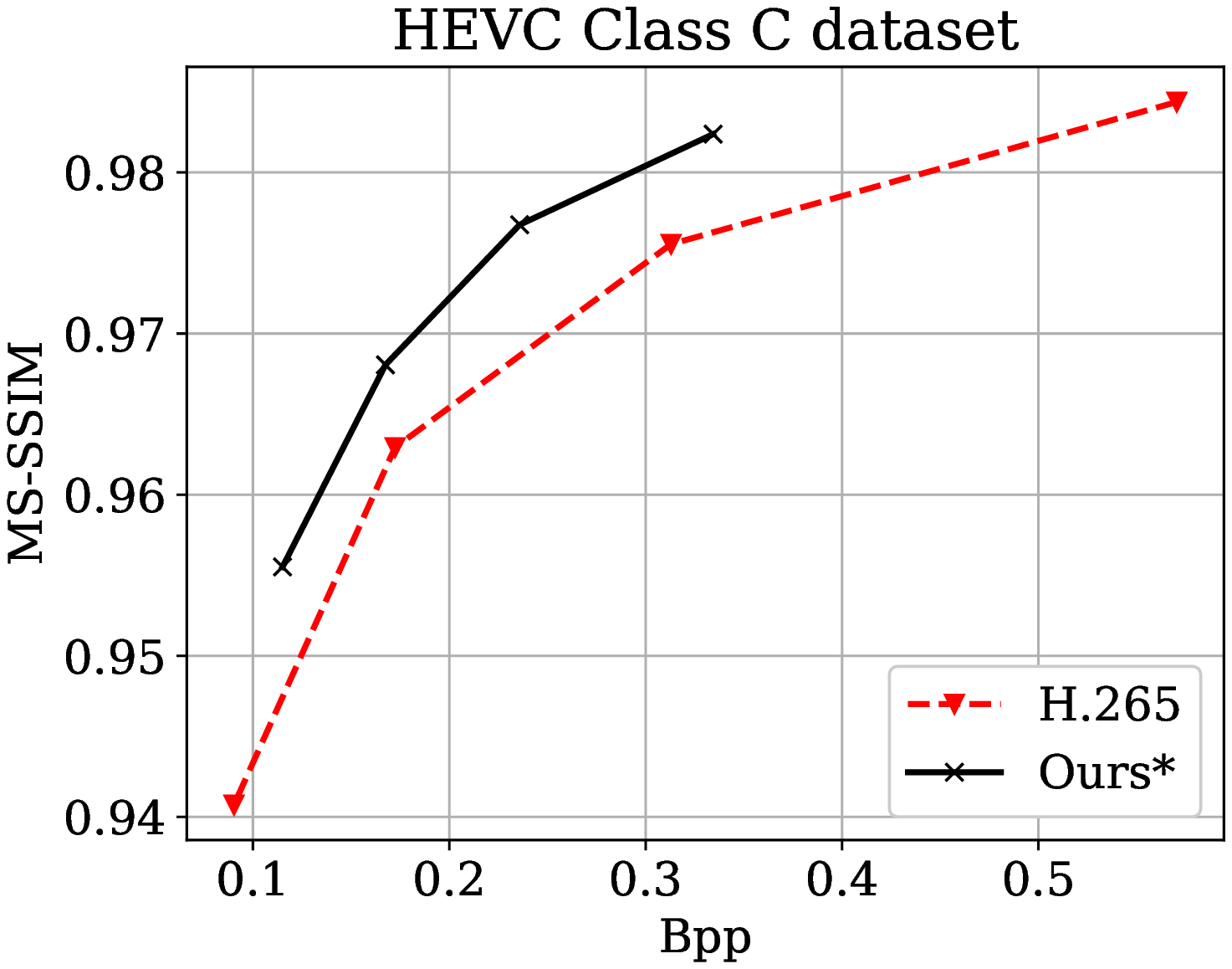}
  \end{minipage}
  \begin{minipage}[c]{0.5\textwidth}
    \centering
    \includegraphics[height=1.75in]{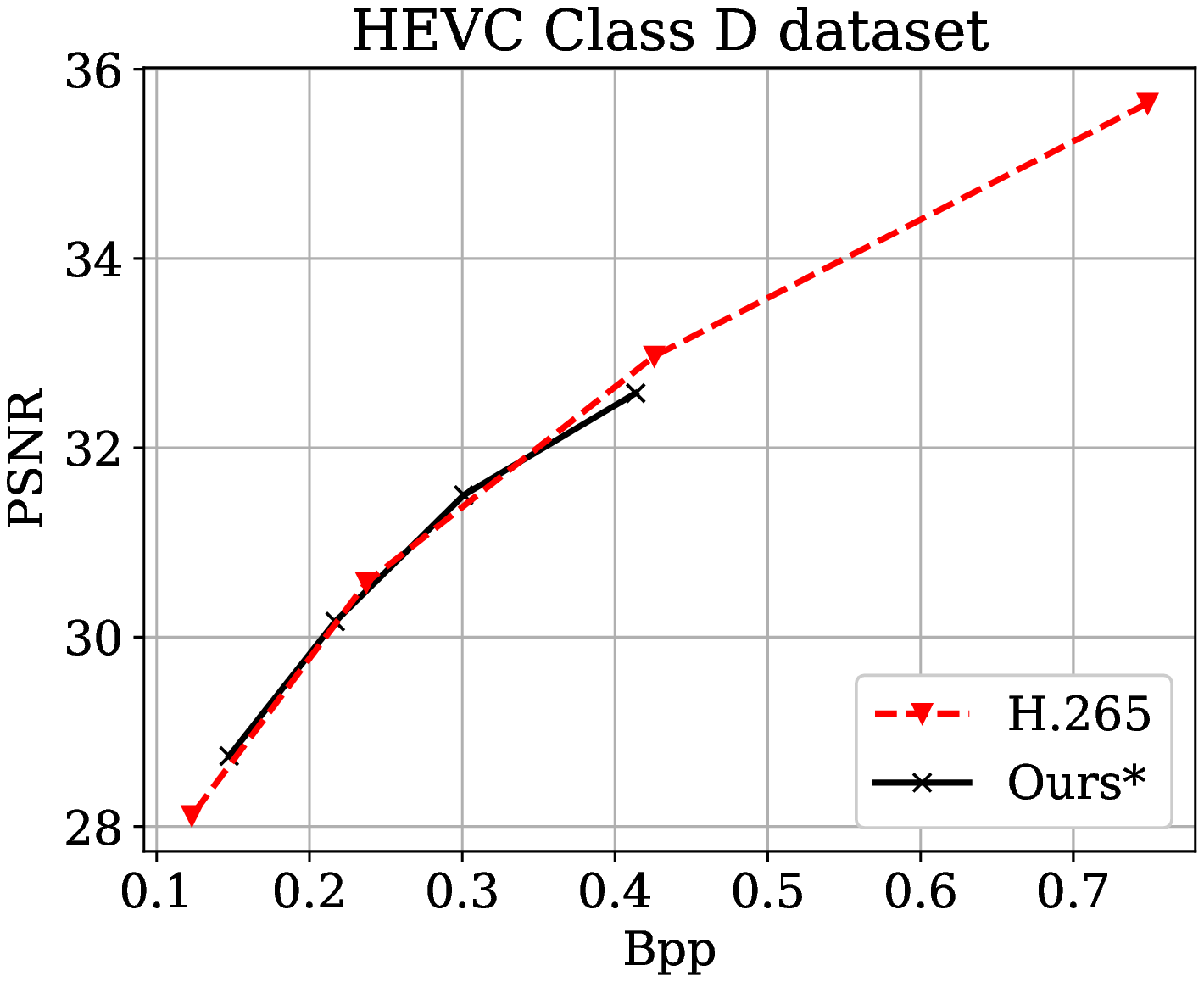}
  \end{minipage}%
  \begin{minipage}[c]{0.5\textwidth}
    \centering
    \includegraphics[height=1.75in]{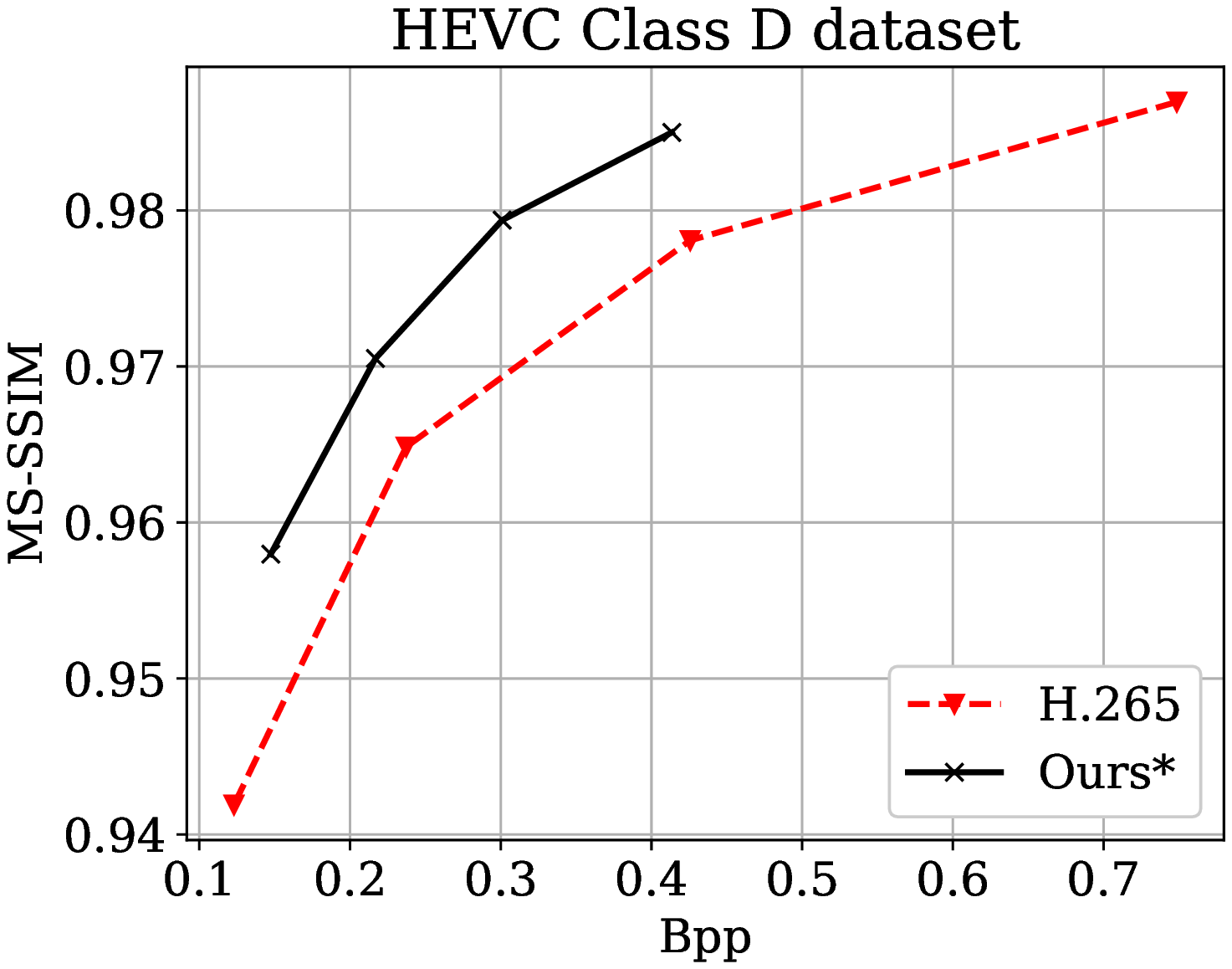}
  \end{minipage}
    \caption{The experimental results of our method and H.265 using variable GoP sizes on the HEVC Class B, Class C and Class D datasets.}
  \label{fig:gop}
\end{figure}

In the previous works like \cite{abdelaziz2019neural,lu2019dvc,wu2018video}, the fixed GoP size is always used for fair comparison. In Figure~\ref{fig:gop}, we provide the results of our method and H.265 when using variable GoP sizes on the whole sequences from the HEVC Class B, Class C and Class D datasets. It is clear that our approach still achieves competitive results when compared with H.265 in terms of PSNR and outperforms H.265 in terms of MS-SSIM.

For H.265 with variable GoP sizes, the command line is provided as follows, 

\textit{ffmpeg -pix\_fmt yuv420p -s WxH -r FR -i A.yuv -c:v libx265 -tune zerolatency -x265-params ``qp=Q:verbose=1" output.mkv}

where \textit{FR, Q} represent the frame rate and the quality. \textit{Q} is set as 22, 26, 30, 34.

\end{document}